\documentclass[journal]{IEEEtran}
\ifCLASSINFOpdf
\else
\fi
\usepackage{graphicx}
\usepackage{url}
\usepackage[hidelinks]{hyperref}
\newcommand{\eqnref}[1]{Eq.~(\ref{#1})}

\newcommand{\secref}[1]{Sec.~\ref{#1}}
\newcommand{\tableref}[1]{Table~\ref{#1}}
\newcommand{\figref}[1]{Fig.~\ref{#1}} 
\usepackage[ruled,linesnumbered]{algorithm2e}
\usepackage{subfigure}
\usepackage{amsmath}
\usepackage{bbm}
\usepackage{amssymb}
\usepackage{booktabs}
\usepackage{multirow}
\usepackage[numbers]{natbib}
\usepackage{xcolor}
\begin{document}
%
\title{GOF-TTE: \underline{G}enerative \underline{O}nline \underline{F}ederated Learning Framework for \underline{T}ravel \underline{T}ime \underline{E}stimation }
%
%
%

\author{
    Zhiwen~Zhang\IEEEauthorrefmark{1},
    Hongjun~Wang\IEEEauthorrefmark{1},
	Jiyuan~Chen, 
	Zipei~Fan\IEEEauthorrefmark{2}, 
	Xuan Song\IEEEauthorrefmark{2},
	and
	Ryosuke	Shibasaki
	\IEEEcompsocitemizethanks{
    \IEEEcompsocthanksitem Hongjun Wang, Zhiwen Zhang,  Jinyu Chen, Zipei Fan, Xuan Song and Ryosuke Shibasaki are with the SUSTech-UTokyo Joint Research Center on Super Smart City, Department of Computer Science and Engineering and Research Institute of Trustworthy Autonomous Systems, Southern University of Science and Technology (SUSTech), Shenzhen, China.  E-mail: \{wanghj2020,11811810\}@mail.sustech.edu.cn; zhangzhiwen@csis.u-tokyo.ac.jp, fanzipei@iis.u-tokyo.ac.jp and songx@sustech.edu.cn.
    \IEEEcompsocthanksitem Zipei Fan, Zhiwen Zhang, Xuan Song, and Ryosuke Shibasaki are also with The University of Tokyo, 5-1-5 Kashiwanoha, Kashiwa-shi, Chiba, 277-8561, Japan; emails: fanzipei@iis.u-tokyo.ac.jp; zhangzhiwen@csis.u-tokyo.ac.jp; songx@sustech.edu.cn and shiba@csis.u-tokyo.ac.jp 
	\IEEEcompsocthanksitem \IEEEauthorrefmark{2} Corresponding to 	Zipei Fan;
	\IEEEcompsocthanksitem \IEEEauthorrefmark{1} Zhiwen Zhang, Hongjun Wang equal contribution;
	}
}
%
%

\markboth{Journal of \LaTeX\ Class Files,~Vol.~14, No.~8, August~2015}%
{Shell \MakeLowercase{\textit{et al.}}: \title{GOF-TTE: \underline{G}enerative \underline{O}nline \underline{F}ederated Learning Framework for \underline{T}ravel \underline{T}ime \underline{E}stimation }}
%



\maketitle

\begin{abstract}
\textcolor{black}{Estimating the travel time of a path is an essential topic for intelligent transportation systems. It serves as the foundation for real-world applications, such as traffic monitoring, route planning, and taxi dispatching. However, building a model for such a data-driven task requires a large amount of users' travel information, which directly relates to their privacy and thus is less likely to be shared. The non-Independent and Identically Distributed (non-IID) trajectory data across data owners also make a predictive model extremely challenging to be personalized if we directly apply federated learning. Finally, previous work on travel time estimation does not consider the real-time traffic state of roads, which we argue can significantly influence the prediction.  To address the above challenges, we introduce GOF-TTE for the mobile user group, \underline{\textbf{G}}enerative \underline{\textbf{O}}nline \underline{\textbf{F}}ederated Learning Framework for \underline{\textbf{T}}ravel \underline{\textbf{T}}ime \underline{\textbf{E}}stimation, which 
I) utilizes the federated learning approach, allowing private data to be kept on client devices while training, 
and designs the global model as an online generative model shared by all clients to infer the real-time road traffic state.
II) apart from sharing a base model at the server, adapts a fine-tuned personalized model for every client to study their personal driving habits, making up for the residual error made by localized global model prediction.}
We also employ a simple privacy attack to our framework and implement the differential privacy mechanism to further guarantee privacy safety. 
Finally, we conduct experiments on two real-world public taxi datasets of DiDi Chengdu and Xi'an. The experimental results demonstrate the effectiveness of our proposed framework.
\end{abstract}


%
\IEEEpeerreviewmaketitle

\section{Introduction}\label{sec:intros}

With the popularization of mobile devices and the maturity of real-time positioning technology, the trajectories of people and vehicles are frequently recorded, resulting in a large amount of position-based data which mainly comes from mobile phones \cite{calabrese2011interplay}, public bicycles \cite{lathia2011smart}, public transportation cards \cite{froehlich2009sensing} and taxis \cite{pan2012land}.
These data is taken from snippets of everyday life, reflecting both individual behavior and city dynamics.
Recently the emerging idea of utilizing these data to promote the construction of smart city draws considerable attention, among which the modeling of travel time estimation (TTE) serves as a key building block. TTE is the fundamental
problem in traffic monitoring, route planning, navigation and taxi dispatching. Having an accurate prediction of the travel time for a given path not only provides convenience to the individual in travel scheduling, but also helps the traffic control center in alleviating road congestion.  

However, although the problem has been widely studied in the past, ranging from Bayesian inference \cite{mil2018modified}, Markov chains \cite{yeon2008travel}, to deep learning models \cite{ide2011trajectory,wang2018will}, they require the travel data from all users which, may contain sensitive identifiable information such as driving habits and locations, to be collected and then trained a model in a centralized way. These conventional methods are faced with severe privacy challenges with the increasing awareness of data security, since one must take the risk of privacy leakage when sharing his own travel information to public organizations.

Fortunately, recent progress in federated learning  \cite{mcmahan2017communication} has shed light on such privacy issues. The fundamental idea of federated learning is to keep user's data locally instead of sharing while collaboratively training a global model by exchanging only training parameters (i.e., model weights). More specifically, the centralized learning mechanism aggregates all clients' data to the cloud sever and trains a centralized model before different clients download it. On the contrary, \figref{fig:tutorial} (a) reveals the federated learning policy, which is proposed to train a local model with local data independently at each client/device, and then sends their model weights to the server where the
updates from different clients are integrated into the global model. The global model later in turn sent its weights to update all local model for each device/client. Ideally, the global model obtained from federated learning has similar or better results than the model obtained from centralized data training on a central server \cite{li2020federated}.

\begin{figure}[t]
	\centering
	\setlength{\belowcaptionskip}{-0.5cm}
	\includegraphics[width=1\linewidth]{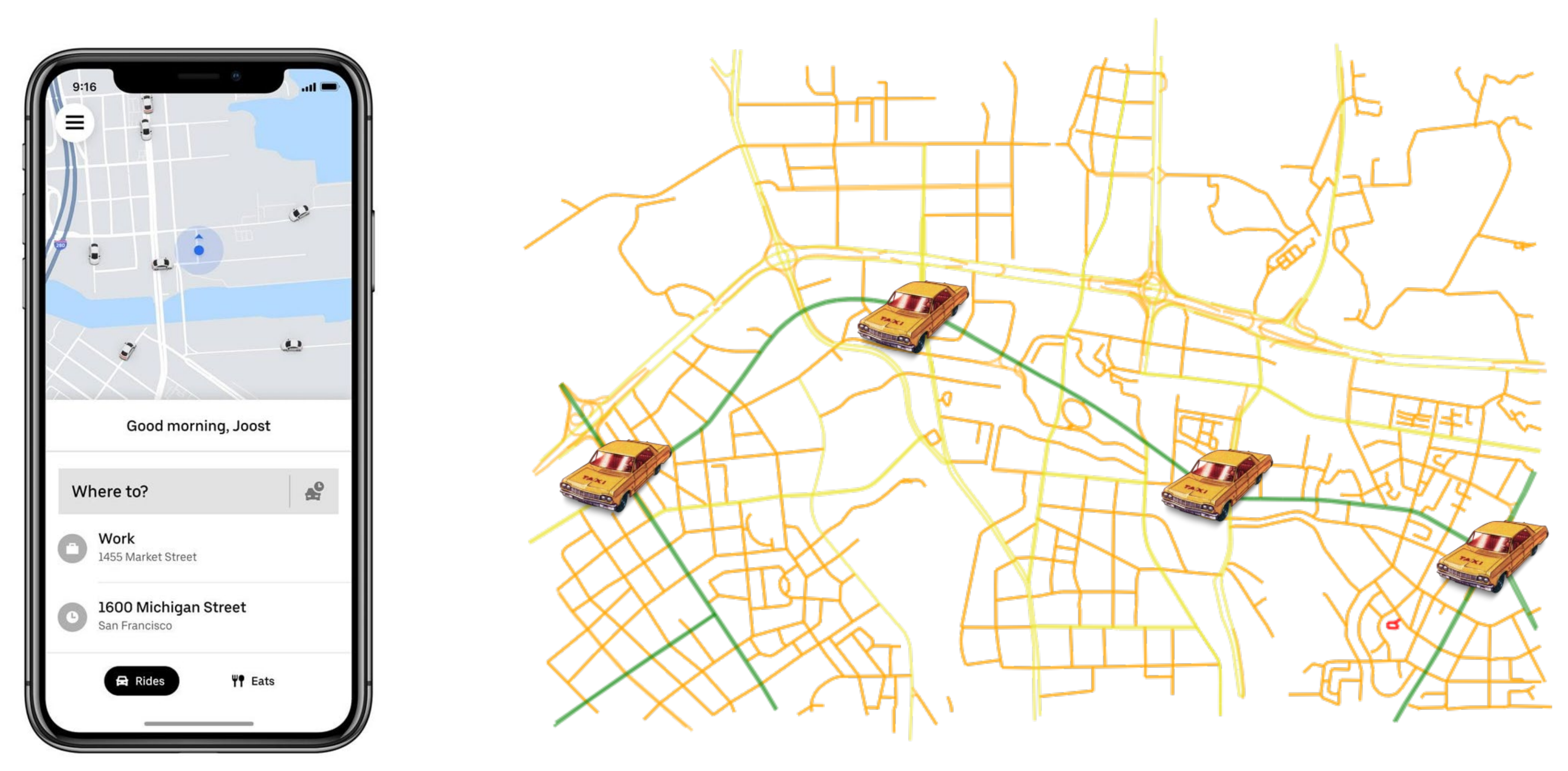}
	\caption{The application of global traffic state in real life. The traffic state is shown in different colors where 1) red - very congested 2) yellow - congested 3) orange - slow, and 4) green - unblocked. A comprehensive perception of the global traffic state helps a lot when planning a route or estimating travel time. }\label{fig:case}
\end{figure}

Learning from non-IID data is one of the major challenges
of federated learning, particularly in the case of TTE, since the travel styles and active areas vary from person to person, resulting in a highly non-IID dataset for each individual user.
While the centralized learning procedure shuffles and mixes the entire dataset to draw batches that hold almost the same data distribution, federated learning suffers when local data is non-IID \cite{zhao2018federated}. Therefore, the global model trained through federated learning might not perform well for each client in the presence of statistical data heterogeneity (i.e., the non-IID data from different clients). We adapt the idea from  \cite{tan2021towards,ke2017lightgbm} to train an extra personalized model for each client to study his/her personal driving habit (i.e., encodes it as latent representation). The personalized model is trained towards the residual error between ground truth and the prediction from the global model. With the cooperation of both a global and personalized model, the final result can be more "personal" and thus more accurate.

Moreover, as mentioned by \cite{li2019learning}, travel time relies heavily on two parts: 1) the statically spatial features of the city road graph (e.g. number of traffic lights, number of lanes, and speed limit), and 2) the dynamically changing real-time traffic condition (i.e., the traffic conditions at different time slots
are temporally-variant due to many random variables such as the daily activities of people, road works, and accidents). \textit{Li et al.,} \cite{li2019learning} proposed to generate a dynamic grid-based global traffic state by combining these heterogeneous factors (both spatial and temporal) together to improve the prediction accuracy. Nevertheless, we argue that it has two important drawbacks: 1) the grid-based traffic state suffers from its coarse-grained characteristics, lacking the modeling for the state of roads, where the travel really happens. 2) its real-time traffic state is calculated manually using time slots and  sub-trajectories rather than using machine learning models to generate, so it is less likely to contain the hidden and high-level traffic information such as traffic accidents. Instead of introducing another new module, we directly design our global federated learning model to be an online generative model that generates the real-time road-based traffic state tensor, inferring the estimated current travel time for each road segment and road intersection. We believe that the real-time traffic state can provide us with more timely traffic information that can have a significant improvement in our prediction. Specifically, we design a spatio-temporal cross product to produce the traffic state by fusing the spatial representations (both road segments and intersections)  and the temporal embedding representations.

Additionally, in the application, the global traffic state addresses even more importance in the field of the intelligent transportation system. It contributes to traffic control, route planning, vehicle scheduling, and other tasks, and plays an important role in alleviating traffic congestion and ensuring public transportation safety. Recently, many companies, such as DiDi \cite{sun2021constructing,liu2019building}, Baidu \cite{fang2020constgat,fang2021ssml} and Amap \cite{dai2020hybrid,gaode.org}, have applied various deep learning methods to traffic state prediction, including studying global traffic states from myriad trajectories \cite{li2019learning}. Thereby, our generative design in federated learning satisfies the interests of both the service provider and individuals, allowing the former to monitor the global traffic state without invading the personal privacy of city residents and providing the latter with convenience. \figref{fig:case} shows one famous case from the view of both the government and individuals. For the government, the traffic state  improves service efficiency, such as ride-sharing services, taxi scheduling, and TTE. For people, the traffic state announces the urban communication efficiency in real-time, which provides tips for the personal daily trip plan.

Summing all mentioned work together, we name our framework as GOF-TTE, \underline{\textbf{G}}enerative \underline{\textbf{O}}nline \underline{\textbf{F}}ederated Learning Framework for \underline{\textbf{T}}ravel \underline{\textbf{T}}ime \underline{\textbf{E}}stimation. The overall structure is shown in \figref{fig:tutorial} (b) and the detailed global and personalized model design is shown in \figref{fig:model}. The contribution of this paper can be summarized as follows:
\begin{itemize}
	\item[$\bullet$] To the best of our knowledge, we are the first to design an online generative federated learning system to estimate the travel time. The online system can provide a more accurate estimation because of its real-time perception of the global traffic state. The traffic state generated via federated learning captures high-level traffic information that companies can collect without privacy leakage. 
	\item[$\bullet$]  Considering non-IID caused by personal driving habits and the inconsistency with data among clients, we propose a personalized federated learning strategy, which reserves a fine-tuned personalized model in the client's device and shares a global model among clients.
	\item[$\bullet$] A practical privacy attack has been given in the travel time estimation task, and the differential privacy technology has been applied when the clients upload his weights to servers. 
\end{itemize}

\textcolor{black}{
The remainder of our paper is summarized as follows: \secref{sec:notations} introduces several basic conceptions (e.g., road Network, route, differential privacy) for our proposed GOF-TTE. \secref{sec:frame} provides our proposed federated framework and data preprocessing. \secref{sec:method} describes our model design and the privacy-preserving technique in detail. \secref{sec:exp_} conducts extensive experiments to evaluate our proposed framework. 
\secref{sec:related} systematically discusses the related work in TTE algorithms, federated learning architecture, and privacy-preserving methods.  \secref{sec:conclude} finally summaries the conclusion and discusses the future works.}

\section{Preliminaries}\label{sec:prelim}
In this section, we will briefly introduce the definition of urban road network, route, travel time estimation problem and differential privacy technology.

\textbf{Definition 2.1 Road Network}. A road network is a directed graph $\mathcal{G}=(\mathcal{V},\mathcal{E},\pi_v,\pi_e)$, where $\mathcal{V}$ denotes the set of nodes (i.e. road intersections) and $\mathcal{E} \subseteq \mathcal{V}\times \mathcal{V} $ is the set of directed edges/links (i.e. road segments). $\pi$ serves as a feature function that can be applied to both $v \in \mathcal{V}$ and $e \in \mathcal{E}$. $\pi_v(v)$ denotes the features of $v$, such as its speed limits and whether it has a traffic light. $\pi_e(e)$ denotes the features of $e$, for example, its road types, road length and the number of road lanes.

\textbf{Definition 2.2 Route}. A route $r_i: e_{1} \rightarrow v_1 \rightarrow e_{2} \rightarrow \cdots \rightarrow v_{|r_i|-1} \rightarrow e_{|r_i|}$ is an alternating sequence with links and nodes.

\textbf{Definition 2.3 Travel Time Estimation (TTE)}. Given a route $r_i$, the task of travel time estimation is to infer the overall traveling time $\Tilde{y}$ necessary for the route in the city.

\textbf{Definition 2.4 (Differential Privacy \cite{dwork2006differential})}  A randomized mechanism $\mathcal{M}: \mathcal{D} \rightarrow \mathcal{R}$ with domain $\mathcal{D}$ and range $\mathcal{R}$ satisfies $(\varepsilon, \delta)$-differential privacy if for any two adjacent inputs $d, d^{\prime} \in \mathcal{D}$ and  any subset of outputs $S \subseteq \mathcal{R}$, there holds
$$
\operatorname{Pr}[\mathcal{M}(d) \in S] \leq e^{\varepsilon} \operatorname{Pr}\left[\mathcal{M}\left(d^{\prime}\right) \in S\right]+\delta,
$$
\textcolor{black}{where $\operatorname{Pr}$ denotes the probability space that is over the coin flips of the mechanism $\mathcal{M}$. Intuitively, differential privacy-based noise is added to the query results so that an attacker cannot tell whether a particular text is in $\mathcal{D}$. In addition, we say that $\mathcal{M}$ is $\varepsilon$-differentially private if $\delta=0$.}



\label{sec:notations}
\section{Framework Design}
\label{sec:frame}
\subsection{Framework Overview}
\begin{figure*}[!tbh]
    \centering
  
    \subfigure[Conventional federated learning framework.]{
        \label{figure:62b}
        \includegraphics[width=0.45
        \textwidth]{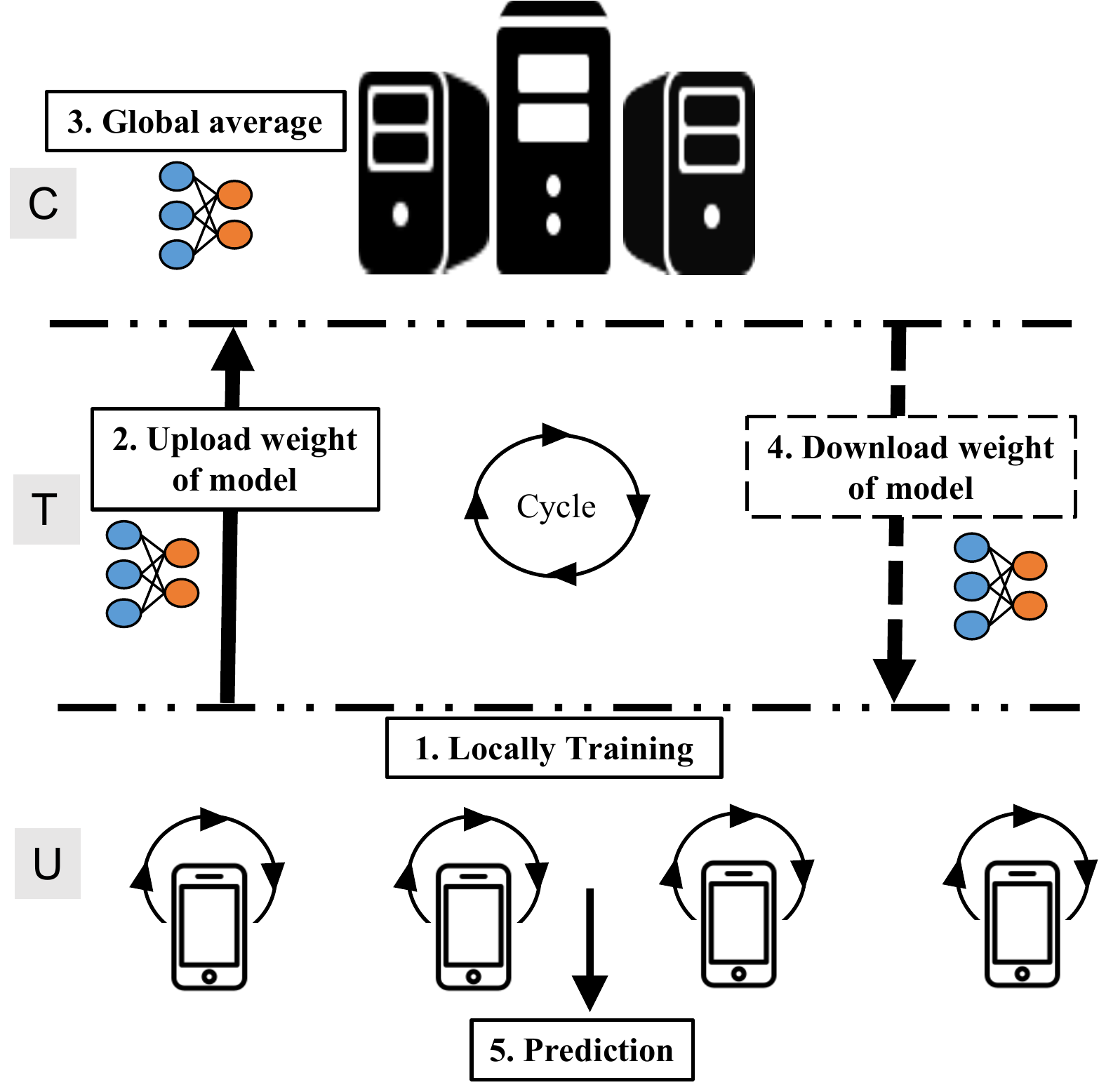}
      }  
      \subfigure[Our proposed federated framework.]{
        \label{figure:61a}
        \includegraphics[width=0.45\textwidth]{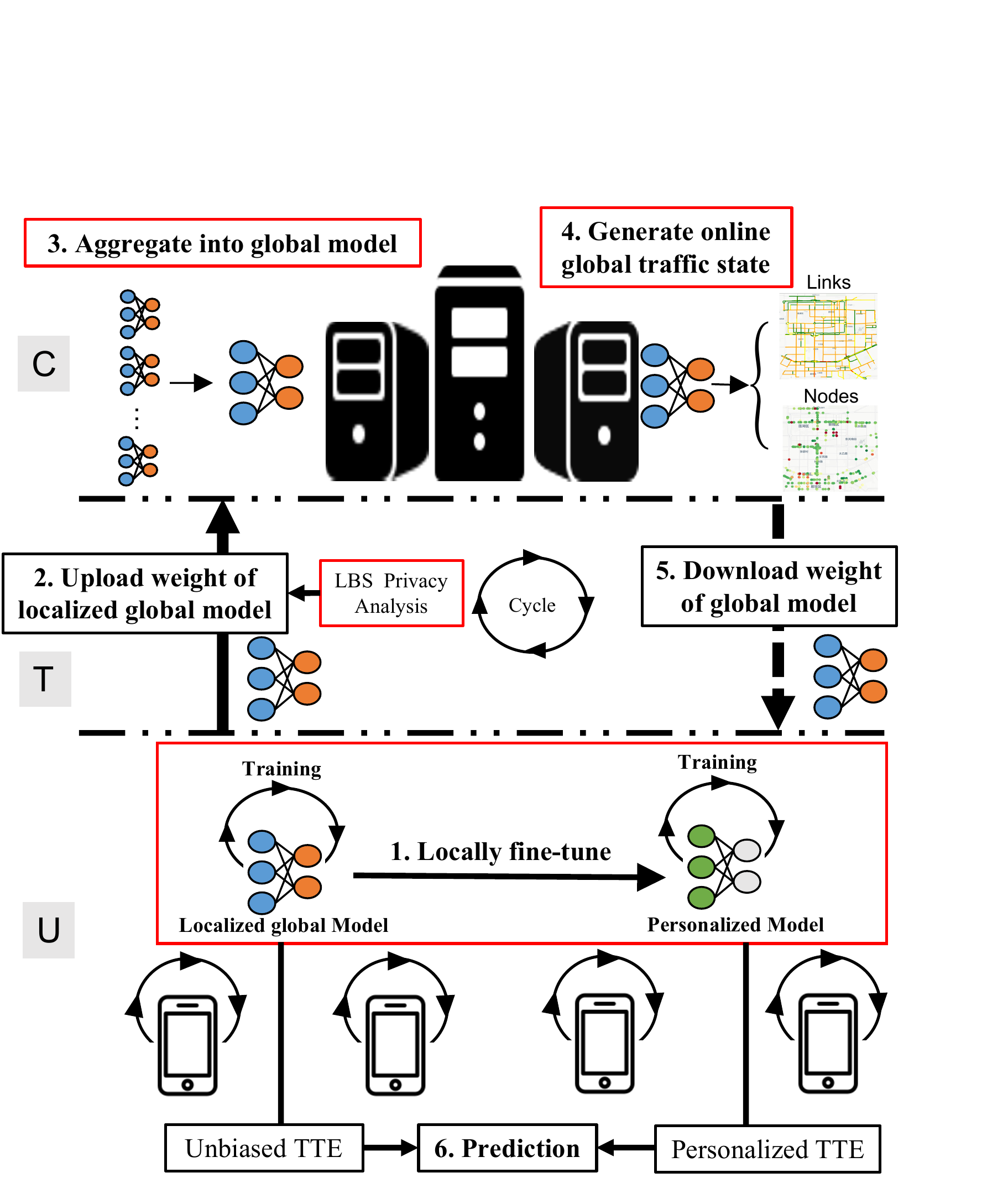}
    
    }
    \caption{Workflow comparison between the directed federated learning system and our proposed framework. "C", "T" and "U" in figure denote the cloud server, transfer environment, and mobile devices (clients) respectively. }
    \label{fig:tutorial}
\end{figure*}
\textcolor{black}{Under the background of the contradiction between the data island phenomenon and data fusion demand, federated learning has been proposed to train the model without sharing data but only exchanging training parameters in the middle stage. Ideally, the shared model obtained from federated learning has similar or better results than the model got from centralized data training on a central server. We here compare the differences between the classical federated system and our proposed framework. The  direct deployment of a federated TTE framework (\figref{fig:tutorial} (a)) for map providers (such as Amap and Google) will lead to several challenges listed as follows:}
\begin{itemize}
	\item[ 1.] \textcolor{black}{In travel time estimation, due to different driving habits and preferences of each driver, the privacy data among drivers has the non-IID characteristics, so the local trajectory data distribution for each user can vary with the user’s locations and preferences.}
	\item[ 2.] \textcolor{black}{As we mentioned, the global traffic state plays an important role for both the problem of estimating travel time \cite{james2021citywide,li2019learning} and the companies interests. How to learn the traffic road state from the individual remains a challenge.}
	\item[ 3.] \textcolor{black}{The direct upload and aggregation of the localized global model parameters may also leak the individual private information in the TTE task.}
	
\end{itemize}

Our architecture (\figref{fig:tutorial} (b)) can ensure that all participants will not disclose personal information to the server, and the server is only responsible for the secure aggregation of local models' parameters, and sends the results to all clients, ensuring that the large family can share a joint trained model.
Repeating this process until the loss function converges completes the training of the entire joint model. As shown in \figref{fig:tutorial} (b), the detailed descriptions of this system are listed as follows:
\begin{itemize}
	\item[ 1.] Each client first trains the localized global model with local data, and then trains the personalized model towards residual error. 
	\item[ 2.] The clients upload their localized global model's parameters with differential privacy to server.
	\item[ 3.] The server performs security aggregation.
	\item[ 4.] The companies/organizations obtained the current global traffic state from the global model.
	\item[ 5.] The server sends the results of the aggregation back to the clients.
	\item[ 6.] Participants update their localized global models and repeat Step 1.
\end{itemize}

\subsection{Data Preprocessing}\label{sec:data_pre}

\subsubsection{Taxi Trajectories.}
 We use two public taxi trajectory datasets from the DiDi Express platform. Xi'an dataset was generated by about 18,000 online car hails in October 2016 in Xi'an, China. Moreover, the Chengdu dataset comprises 9 737 557 trajectories of 14864 taxis in August 2014 in Chengdu, China. The GPS points of both datasets have been mapped to roads via a map-matching tool FMM \cite{yang2018fast} and the time interval of sampling GPS points is 2-4s, ensuring that the vehicle trajectories can correspond to the actual road information.

\subsubsection{Map Gridding} We employ standard quantization to divide the urban map into multiple uniform sub-regions to understand the driver's preferences. Formally, we divide the geographical space of the city into multiple areas that do not intersect.

\subsubsection{Road Network Feature Extraction}  This paper uses two road networks: Chengdu Road Network and Xi'an Road Network. Both of them are extracted from OpenStreetMap. Xi'an road network contains 4780 edges (i.e., road segments) and 3832 nodes (i.e., road intersections), and the Chengdu road network contains 8221 edges and 5182 nodes. To take care of the privacy-preserving concerns and to generate an unbiased traffic state, we here extract the urban road network features from OpenStreetMap (OSM)\footnote{https://www.openstreetmap.org/}, which is a free, open-source, and editable map service jointly created by the public network. Furthermore, we extract the road segments features from \textit{highway tag}, which are listed as follows:
\begin{itemize}
	\item[$\bullet$] \textit{Road types}: freeway, trunk, primary, 	secondary, etc.;
	
	\item[$\bullet$] \textit{special types }: whether it is bridge, tunnel, footway, etc.;
	
	\item[$\bullet$] \textit{Otherwise features}: speed limits, length of road, width of road , number of lanes, etc. ;
\end{itemize}
and intersections from \textit{node tag}, such as 
\begin{itemize}
	\item[$\bullet$] \textit{Tags of node}: e.g. speed camera, traffic signals, crossing signs, turn circle and bus stop;
	
	\item[$\bullet$] \textit{Junction types}: e.g. X-juction, T-junction and 5-way junction: 
	
	\item[$\bullet$] \textit{Otherwise features}: e.g. unique ID and GPS coordinates;
\end{itemize}

\begin{figure}[t]\setcounter{subfigure}{0}
	\centering    
	\subfigure[Speed limit] {
		\includegraphics[width=0.20\linewidth]{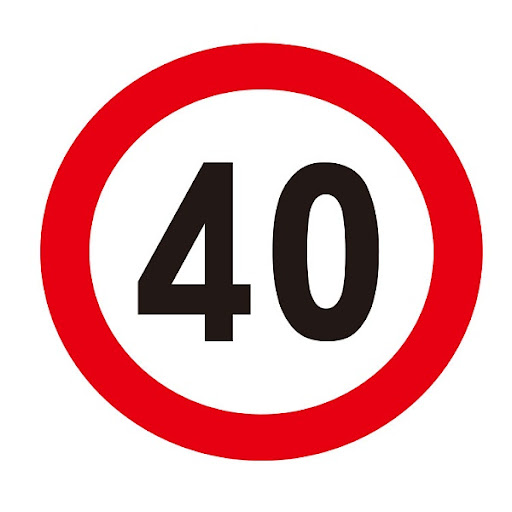}}     
	\hfill
	\subfigure[Crossing] {
		\includegraphics[width=0.20\linewidth]{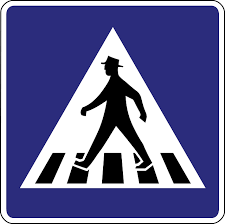}}     
	\hfill
	\subfigure[No right turn] {
		\includegraphics[width=0.20\linewidth]{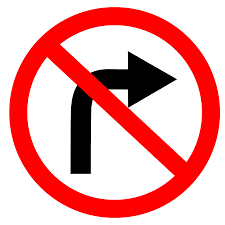}}     
	\hfill
	\subfigure[Traffic light] {
		\includegraphics[width=0.20\linewidth]{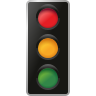}}     
	\caption{The sample cases of \textit{tags of intersection (node)} in OSM.}
\end{figure}

\subsubsection{Drivers' Profile Feature Extraction}\label{sec:drivers_Features}
\textcolor{black}{Most existing literature tries to understand and analyze driving behavior \cite{ezzini2018behind,dong2016characterizing}. However, intuitively, driving behavior is also an essential factor in the problem of TTE. For example, novice drivers are unfamiliar with the urban road network and driving skills. So their expected travel time tends to be longer than experienced drivers. Based on this assumption, we extracted the following drivers' profile features.}

\begin{figure}[!bht]\setcounter{subfigure}{0}
	\centering  
		\subfigure[Average break start time.] { 
	 \label{figure:d1}
	\includegraphics[width=0.48\linewidth]{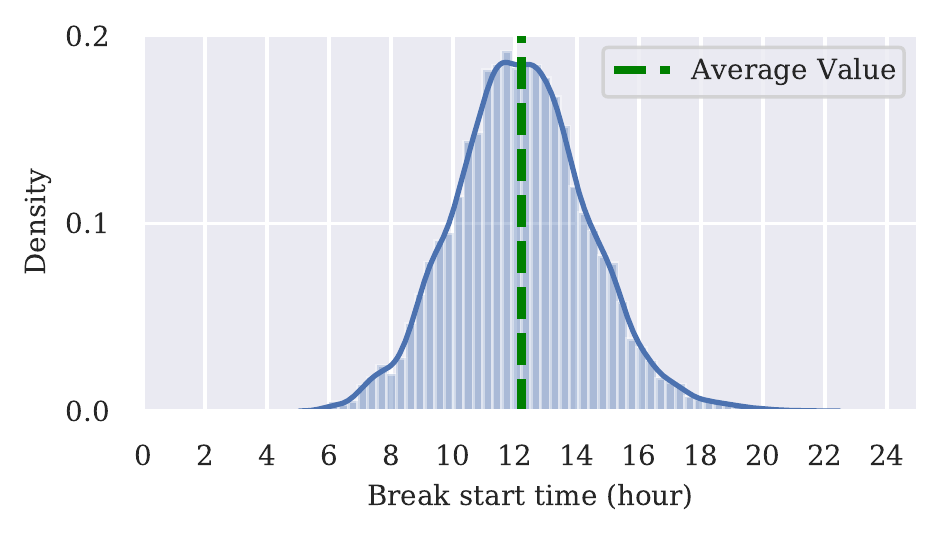}} 
	\subfigure[Average break end time.] {
	 \label{figure:d2}
	\includegraphics[width=0.48\linewidth]{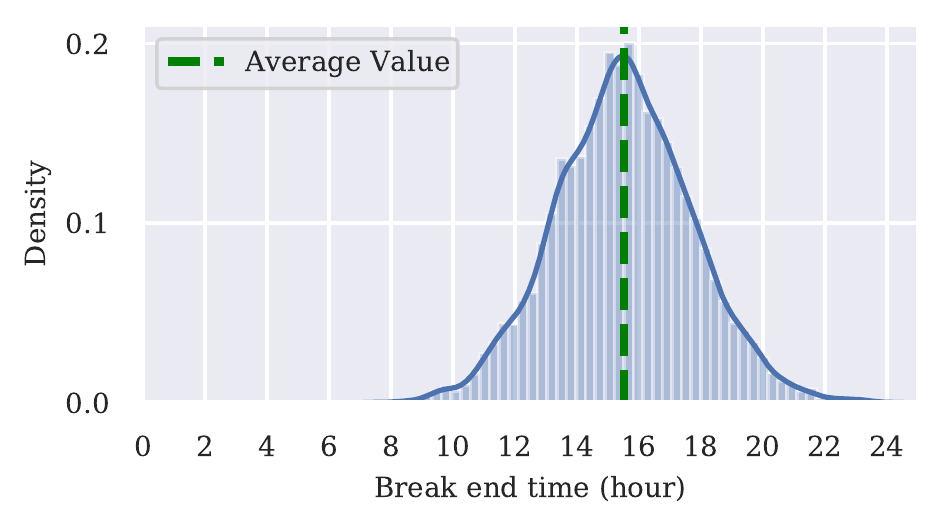}}

	\subfigure[Average driving trip distance.] {
	     \label{figure:d3}
		\includegraphics[width=0.48\linewidth]{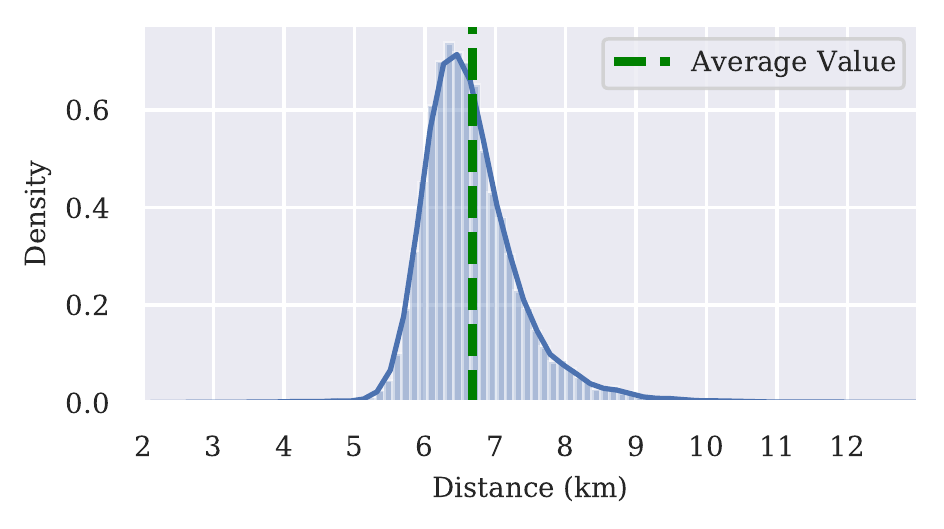}}     
	\subfigure[Number of served trips.] {
	     \label{figure:d4}
		\includegraphics[width=0.48\linewidth]{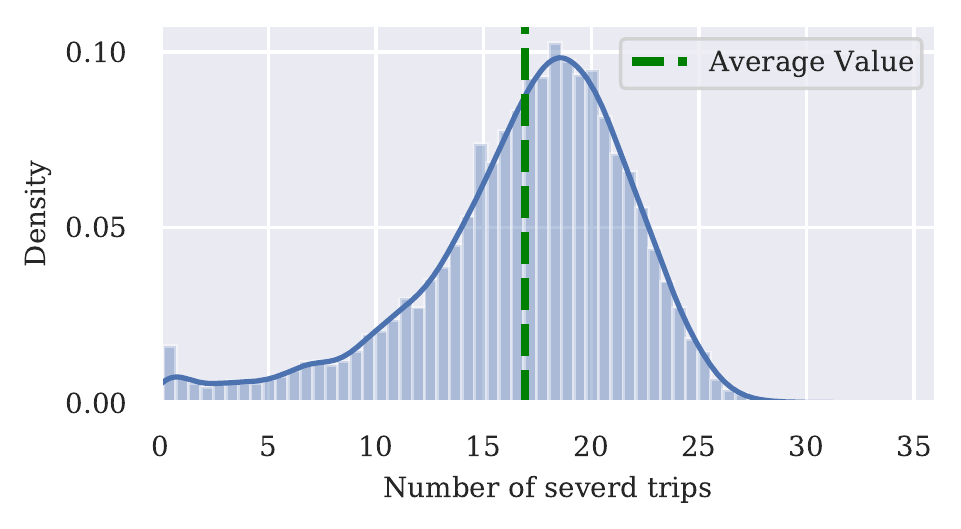}} 
	\caption{\textcolor{black}{Drivers' profile feature analysis.}}
	\label{fig:driver_profiles}     
\end{figure}

\underline{$c_{d,1} \& c_{d,2}:$ Break start \& end time.} \textcolor{black}{Each taxi driver could take a break during work. The driver's preference for break and end time can speculate about his/her usual working time. As a matter of experience, the longer you work, the slower you drive. The distributions of both two features are shown in \figref{figure:d1} and \figref{figure:d2}, respectively. Here, the x-axis is the average break start \& end time for each day.}

\underline{$c_{d,3} \& c_{d,4}:$ Frequently visited regions and roads.}
\textcolor{black}{Different regions and road segments have different road conditions. Therefore, we can speculate on the driver's preference and threshold for congestion. For example, if a driver always visits the downtown region at the morning peak, we guess he/she does not avoid the congestion.}

\underline{$c_{d,5}:$ Average driving distance.} \textcolor{black}{Taxi drivers can have his/her preference for the distances of served trips. Some drivers could prefer to choose the orders for long trips because each trip can earn much more. Other drivers could intend to choose short trips due to the relaxation and convenience. \figref{figure:d3} presents the drivers' average driving distance for each order. From this point, we can know whether one driver prefers the long-distance order or not. Generally, the longer you drive, the slower you go.}

\underline{$c_{d,6}:$ Number of trips served.} \textcolor{black}{Each driver has his/her strategy for looking for passengers. \figref{figure:d4} provides the drivers' number of orders for each day. This is an indicator to distinguish the novice and experienced drivers. In general, the travel time for experienced drivers is shorter.}

\textcolor{black}{From the \figref{fig:driver_profiles}, we can find that taxi drivers obviously have  different driving habits, which could influence their estimation performance.}

\begin{figure*}[t]
	\centering
	\setlength{\belowcaptionskip}{-0.5cm}
	\includegraphics[width=1\linewidth]{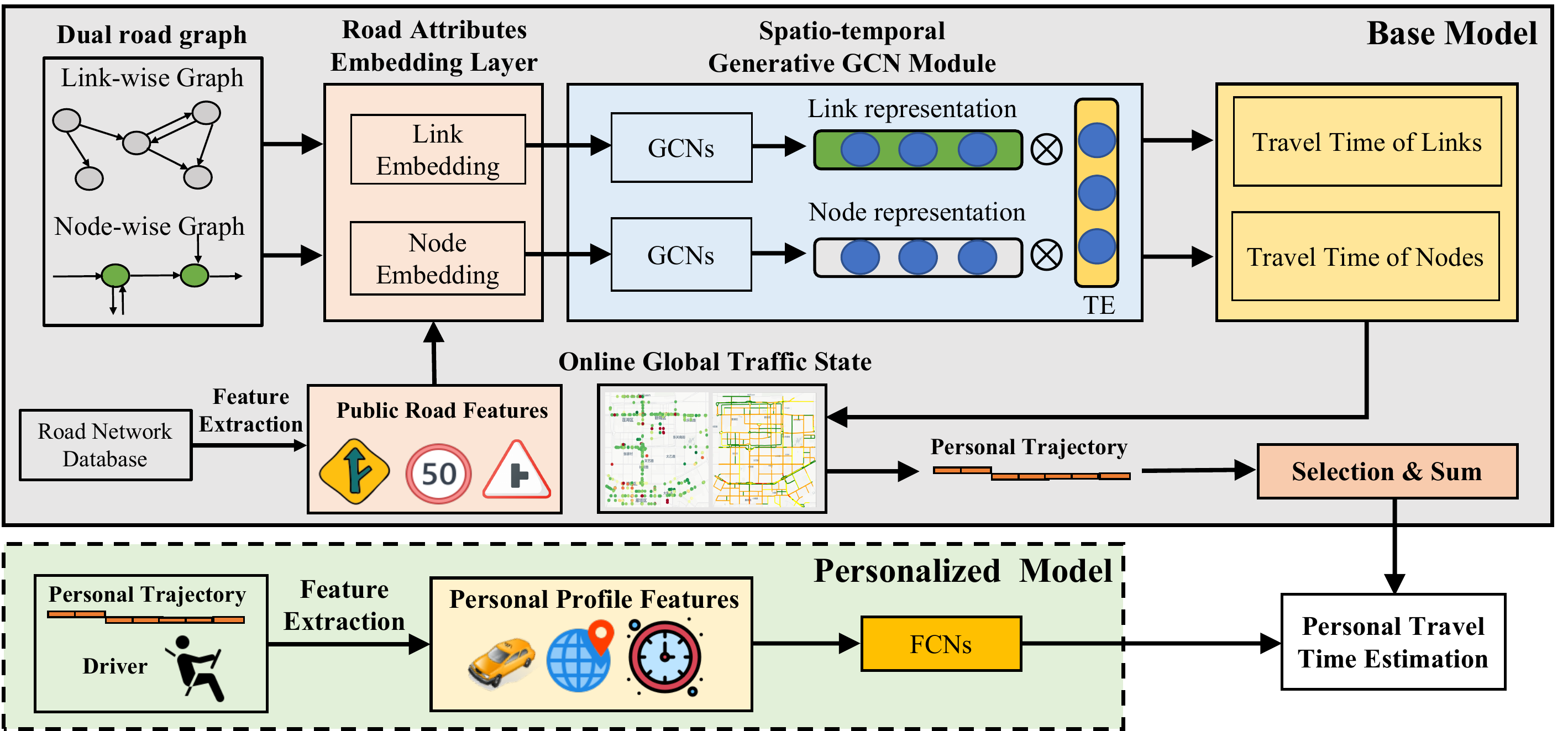}
	\caption{\textcolor{black}{The model design of our federated learning framework that can be divided into two parts: a base model and a personalized model. Here, TE: temporal embedding; FCNs: fully connected networks.}}\label{fig:model}
\end{figure*}

\section{Methodology}
\label{sec:method}
\textcolor{black}{The detailed base and personalized
model design is shown in \figref{fig:model}. The base model is shown at the top of the figure, which generates the online global traffic state for the cloud servers and infers the non-personalized travel time for each client.}

\textcolor{black}{In the subsequent sections, we will introduce our proposed base model, personalized model, privacy-preserving mechanism, and online training procedure.
}

\subsection{Base Model}
\textcolor{black}{In this study, the base model mainly provide the functions of two parts: 1) \textbf{Localized global model}: this is the well-trained version of base model after the local training for each client, which could also provide non-personalized TTE for each client; 2) \textbf{Global model}: the global model is the aggregated model after each round of communication, which aims to generate online global traffic state for cloud server, including both road segment-wise and intersection-wise road network.}

\subsubsection{Road Attributes Embedding Layer}
To accurately estimate the travel time, the spatial feature of the city road network is very important. To get there, we embed both the features of roads segments and road intersections into the model through an embedding layer. \figref{fig:model} shows an overview of this procedure. The input of the layer is a dual road graph, where link-wise graph and node-wise graph characterize the dependencies of road segments and intersections respectively. 

\subsubsection{Spatio-temporal Generative GCN Module}  In this subsection, we will introduce our spatio-temporal generative module, which contains the spatial GCNs layer and a spatio-temporal cross-product layer. The former aims to capture the spatial dependencies of the road network and the latter aims to generate the global state comprised of road segments  and intersections, respectively. 

\textbf{Spatial GCNs Layer.} Recently, a bunch of works on Graph Convolutional Network (GCN) \cite{kipf2016semi}  have been proposed to capture the complex relations in the road network \cite{li2017diffusion}. In this study, we employ the GCN into our framework to process the public features from road network. Formally, given a graph signal  $h^{(l)} \in \mathbb{R}^{N\times d^{(l)}}$, where $N$ denotes the number of samples (links/nodes) and $d^{(l)}$ is the features size in $l^{th}$  layer, a typical formulation of graph convolutional layer with $C$-hop is 
\begin{align}
	G C N_{\mathcal{G}}(h^{(l)} ; W^{(l)}, \theta^{(l)})=\sigma\left(\sum_{c=0}^{C} \theta_{c}^{(l)} L^{c} h^{(l)}\right) W^{(l)}.
\end{align}
In our case, the $h^{(l)}$ stands for the hidden state of (links/nodes) from the output of the  \textit{Road Attributes Embedding Layer}.

\textbf{Spatio-Temporal Cross Product Layer.} 
\textcolor{black}{In fact, the global traffic state varies from time to time (e.g., the changes in non-rushing and rush hour). Meanwhile, the online system requires a time-variant feature representations to update the current traffic state.} To solve this problem, we model the temporal dynamically correlations to combine with the static spatial representations. More specifically, we encode the day-of-week and time-of-day of each time step $t_k$ into $\mathbb{R}^7$ and $\mathbb{R}^\mathcal{K}$ using one hot encoding, and concatenate them with the embedding of HolidayID (holiday or not). \textcolor{black}{With the aforementioned background of TTE in federated learning, we here consider the task of base model is to recover the real traffic state, which can be formulated as the tensor completion problem, in which the two modal components (both spatial and temporal aspects) have been widely modeled and analyzed \cite{jin2021spatial,jepsen2020relational}. Since the urban road traffic state has a typical spatio-temporal distribution characteristics \cite{li2017diffusion}, which can be divided into two levels of intersections and road segments in spatial aspect, and into time-of-day, day-of-week and HolidayID in temporal aspect. The structured time-space data of traffic state can be organized into the form of multidimensional tensor, which lays a foundation for the application of tensor theory in data completion \cite{fan2014cityspectrum}. In addition, a temporal attention mechanism is also employed to capture the relations of adjacent time steps. Let one day be into $K$ time steps by the time interval. Then the model reconstructs the road traffic state by employing the 1st order CP decomposition between the hidden state $h^{(l+1)}$ generated by the GCNs and the temporal embedding $\mathcal{T} \in R^{K\times \mathcal{I}}$, where $\mathcal{I}$ is the size of temporal embedding vector. Thus, we can obtain the output of spatio-temporal cross product layer:}
\begin{align}\label{equ:baseout}
	Y_e &=\left(h_{e}^{(l+1)} W_{\mu_e}+b_{\mu_e}\right) \otimes \text{att}(\mathcal{T}) \nonumber\\
	&=\left(h_{e}^{(l+1)} W_{\mu_e}+b_{\mu_e}\right) \otimes \frac{\exp{(\mathcal{T}_{t_k}})}{\sum_{t_k}^{K}\exp{(\mathcal{T}_{t_k})}} 
\end{align}
\begin{align}\label{equ:baseout}
	Y_v &=\left(h_{v}^{(l+1)} W_{\mu_v}+b_{\mu_v}\right) \otimes \frac{\exp{(\mathcal{T}_{t_k}})}{\sum_{t_k}^{K}\exp{(\mathcal{T}_{t_k})}} 
\end{align}
where $W_{\mu_e},W_{\mu_v} \in \mathbb{R}^{d^{(l+1)}\times \mathcal{J}}$,  $b_{\mu_e} \in \mathbb{R}^{|\mathcal{E}|}$ and $b_{\mu_v} \in \mathbb{R}^{|\mathcal{V}|}$ are the parameters in the fully connected layer to reconstruct the output of spatial GCNs layer. Based on the spatio-temporal cross product of spatial feature vectors $h^{(l+1)}W_{\mu}+b_{\mu}$ (for both nodes and links) and the temporal attention-based vector $\text{att}(\mathcal{T})$, $Y^{t_k}_{e} \in \mathbb{R}^{|\mathcal{E}|}$ and $Y^{t_k}_{v} \in \mathbb{R}^{|\mathcal{V}|}$ are the outputs of the base model and denoted as the current traffic state  of links and nodes at the $t_k$ time step, respectively.

\subsubsection{Localized Global Model Prediction} Since we have obtained the estimation time of road segments and intersections from \eqnref{equ:baseout}, next we will introduce how to generate the non-personalized travel time-based prediction for each client. With the aforementioned definition of the route in \secref{sec:prelim}, a route $r$ in this study is an alternating sequence of links and nodes, which occurs at the time step $t_k$. Therefore, we have the localized global output of this route:
\begin{align}\label{equ:preds}
	\widehat{y}=\sum_{e_i \in r}^{|r|}Y_{e_i}^{t_k}+\sum_{v_j \in r}^{|r|}Y_{v_j}^{t_k}
\end{align}
After local training of the base model based on the trajectory data of  each client, we can acquire the localized global model and then upload this trained model to cloud server, which is used to conduct federated model aggregation.

\subsubsection{Online Global Traffic State Generation}
\begin{figure*}[t]
	\centering
	\setlength{\belowcaptionskip}{-0.5cm}
	\includegraphics[width=0.85\linewidth]{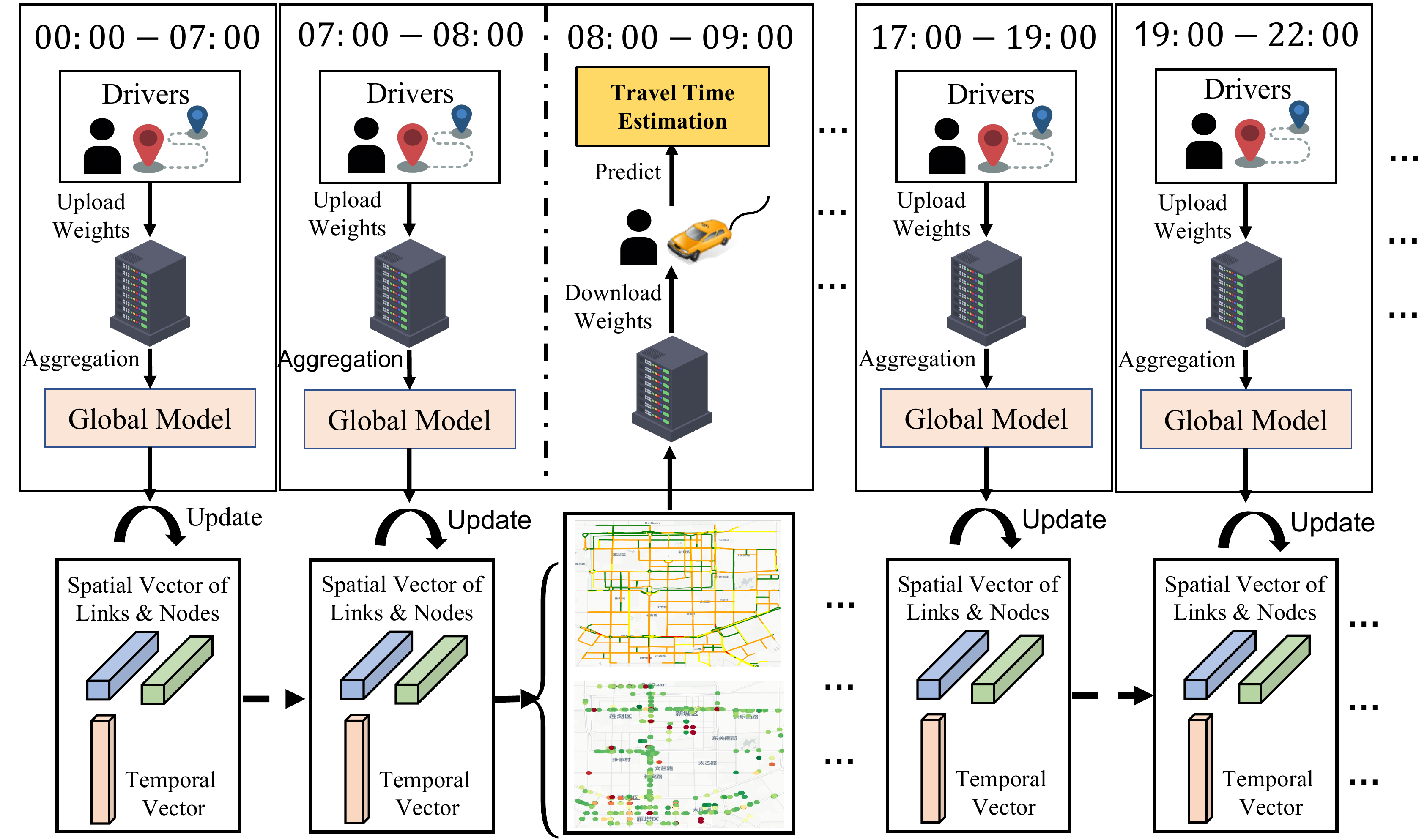}
	\caption{ Online traffic state aggregation process of our proposed system.}\label{fig:training-procedure}
\end{figure*}

\textcolor{black}{  When the cloud server collects weights of the trained localized global model from each user/client, we here aim to generate an online global traffic state based on the aggregated global model. \figref{fig:training-procedure} shows a schematic of our aggregation process for the online global traffic state, which provides an extensible and flexible method to infer the online average travel time for each road segment and road intersection from the road network. Technically, we divide the time of day into five parts: 1) Midnight hours (0:00–7:00); 2) Morning rush hours (7:00–9:00); 3) Non-rush hours (9:00–17:00); 4)  Evening rush hours (17:00–19:00), 5) Evening hours (19:00–24:00). This design could avoid the online aggregation process's high computation and communication cost based on the relatively stable traffic state in the non-rush hours. In our case, let $\Delta$ be a parameter indicating the online aggregation frequency, we set the time interval $\Delta$=0.5 hour during 7:00-9:00 (frequently changing traffic state in the morning peak), $\Delta$=2 hours during 9:00-17:00 (stable traffic state), $\Delta$=0.5 hour during 17:00-19:00 (evening peaks), $\Delta$=2 hours during 19:00-23:00 (stable traffic state) and $\Delta$=4 hours during 23:00-7:00 (sleeping time). Our online learning process for global traffic state can be viewed as a repetitive prediction process, as shown in \figref{fig:training-procedure}. For example, selected users take part in the training based on their trajectories during 07:00-07:30 and upload their trained localized models to the server. When clients send a TTE query for their trajectories during 07:30-08:00, the cloud server will directly provide each client's global traffic state of the last time step for non-personalized estimation through \eqnref{equ:preds}. In addition, our proposed online aggregation also avoids the missing value problem due to the observation uncertainty or limited reports from crowdsourcing systems. Formally, the task of online global traffic state generation can be described as inferring the citywide travel time for each road segment and intersection by acquiring the fusion result $Y_{e} \in \mathbb{R}^{|\mathcal{E}|}$ and $Y_{v} \in \mathbb{R}^{|\mathcal{V}|}$ from the aggregate global model. Relying on the spatio-temporal feature representation of the global model, the server can acquire the complete road condition, which accords with the structure of the road network. }

\subsection{Personalized Model}
Due to the various driving habits of people, the trajectory data collected by different mobile users are very likely to be unevenly distributed. Therefore, many common optimization algorithms with the assumption of IID data are not suitable for our distributed traffic prediction for the mobile user group. If we directly use the localized global model to predict travel time for clients, there will be bias which reduces prediction accuracy \cite{zhao2018federated}. In this work, we introduce a personalized model for each client to study their personal driving habits. We here use the drivers' profile features described in \secref{sec:data_pre}. For these dense features, for example, $f_{d,5}$, $f_{d,6}$, and sparse features, for example, $f_{d,1}$,$f_{d,2}$, we apply a linear transformation and embedding technique, respectively. After embedding the profile features, we concatenate the above hidden states and employ a fully connected layer to produce the personal travel time bias as
\begin{align}
	\widehat{bias}= W_{u}^{(l)} X_{u}^{(l)} + b_{u}^{(l)}
\end{align}
where $X_{u}^{(l)}$ is the hidden state of personal features in $l^{th}$ layer. $W_{u}^{(l)}$ and $b_{u}^{(l)}$ are the parameters of the fully connected layer in $l^{th}$ layer. We consider the final output to be the cooperation between the localized global model and the personalized model. The former produces the non-personalized prediction and the latter makes up for the residual error between the ground truth and the former's output.

\subsection{Privacy-preserving Mechanism}
\label{sec:privacy_sec}

\subsubsection{Privacy Attack}
\begin{figure}[!hbt]
	\centering
	\includegraphics[width=1\linewidth]{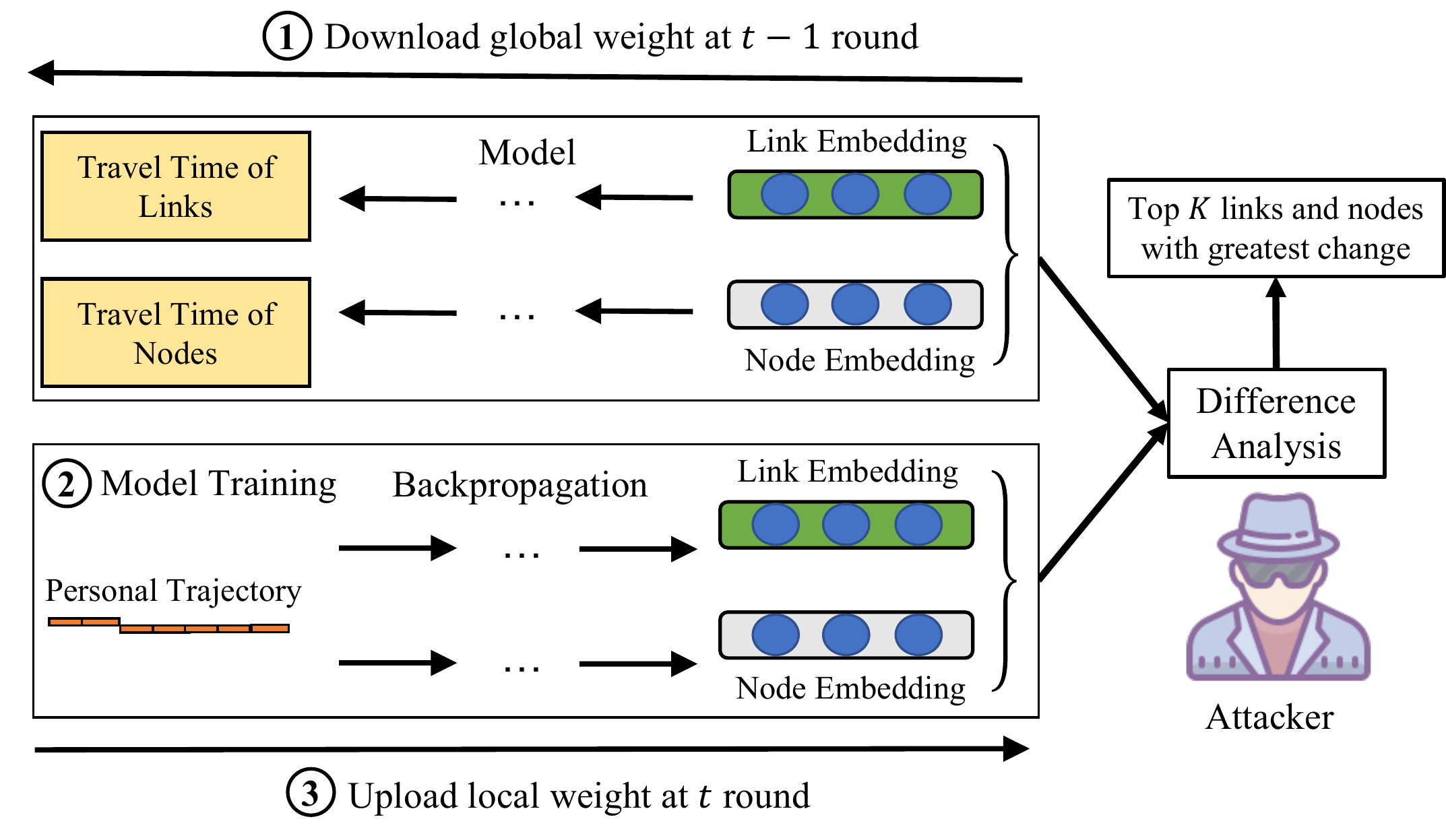}
	\caption{Schematic of privacy attack in base model. } 
	\label{fig:privacy_attack}
\end{figure}

\begin{figure}[h]\setcounter{subfigure}{0}
	\centering    
	\subfigure[Case of Driver 1.] {
		\includegraphics[width=1\linewidth]{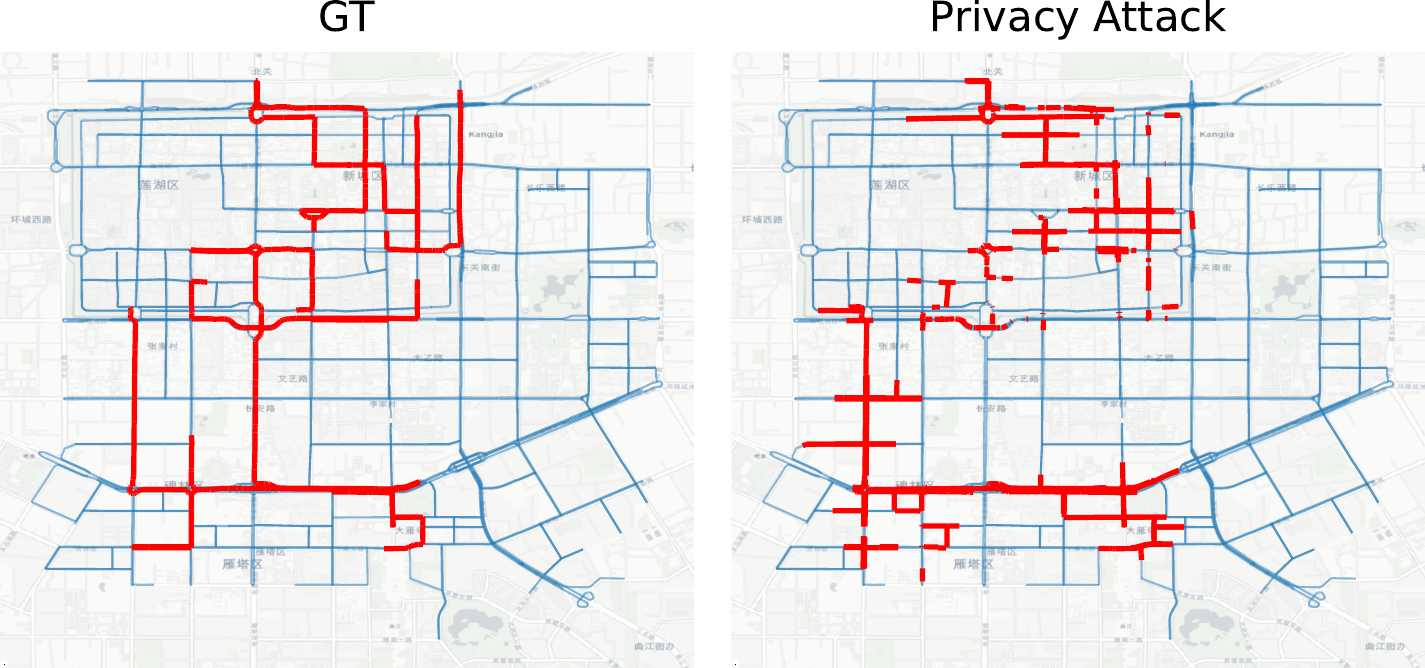}}     
	\hspace{5mm}
	\subfigure[Case of Driver 2.] { 
		\includegraphics[width=1\linewidth]{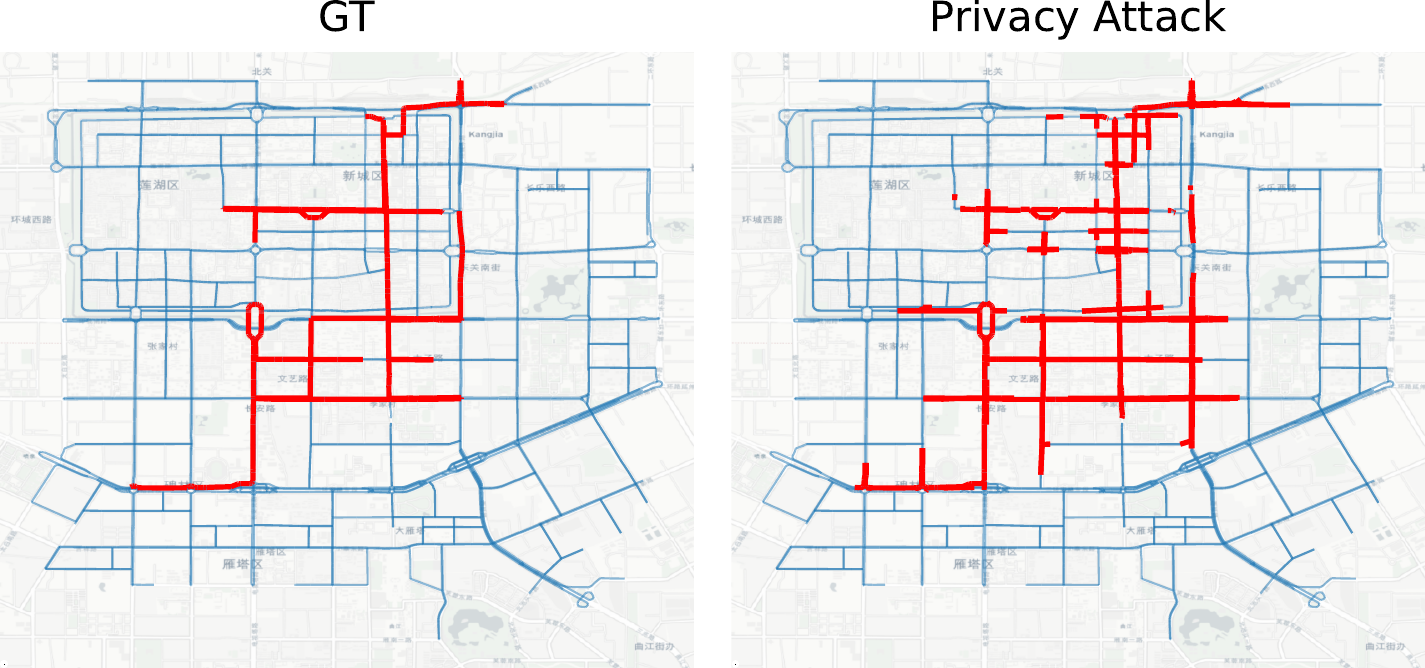}} 
	\caption{The comparisons between the ground-truth road segments and the privacy attack result for two driver cases from Xi'an datasets.}
	\label{fig:privacy_attacks}     
\end{figure}

In distributed machine learning, both the centralized worker-server architecture and the decentralized all-Reduce architecture involve the sharing of local gradients. In the former, worker gradients are visible on the server, while in the latter, neighbor gradients are shared. Recently, a bunch of work \cite{zhu2020deep,geiping2020inverting,zhao2020idlg} has raised people's awareness about the security of gradients. In our case, since the proposed model architecture is composed of a base model and a personalized model, an intuitive privacy attack approach is to compare the differences between the base model's parameters delivered from server at $t-1$ round and current parameters uploaded by client device at $t$ round (\figref{fig:privacy_attack}), which is based on the assumption that the difference between localized global model parameters and the global model parameters is approximately equal to the differences between gradients \cite{zhu2020deep}. \figref{fig:privacy_attacks} represents two cases of privacy attack by using difference attack. We show the top 500 road segments with the largest change in the link embedding layer between the global model and the localized global model. Through the figure, we can see that the privacy attack achieves 52.60\% and 83.50\% overlap respectively with the ground truth. 

\subsubsection{Model Privacy-preserving with Differential Privacy}

Differential privacy \cite{dwork2008differential} defined in \secref{sec:prelim} can solve two defects of traditional privacy protection methods.
First, the differential privacy-based protection method assumes that the attacker can obtain information about all records except the target record, and the sum of this information can be understood as the maximum background knowledge that the attacker can master.
Under this maximum background knowledge assumption, differential privacy-based protection does not need to take into account any possible background knowledge possessed by the attacker, because this background knowledge is unlikely to provide more information than maximum background knowledge.
Second, based on a solid mathematical foundation, differential privacy technology has a strict definition of privacy protection and provides a quantitative evaluation method, which makes the privacy protection level with different parameter settings comparable. In the final, there is no doubt that the privacy protection method could influence the model performance, and we need to achieve a good balance between the privacy protection and model prediction performance.

\textcolor{black}{More specifically, we aim to evaluate the privacy risk based on differential privacy and to analyze the trade-off between model performance and security guarantee. In this work, we construct the noisy data with a differential privacy-based Laplace mechanism, and add them into the weight of the local globalized model when each client uploads their trained models onto a cloud server. The Laplace mechanism preserves $(\varepsilon,0)$-differential privacy, which perturbs each coordinate with noise drawn from the Laplace distribution. }

\subsection{Model Training Procedure}

\label{model_training}
Step by step training in personalized federated learning is a common strategy \cite{guo2021pfl,yu2020salvaging}, which means that it defines the training priorities between the localized global model and the personalized model. In our case, inspired by the federated learning with personalized layers \cite{arivazhagan2019federated}, we first train the base model to learn the non-personalized traffic state for each client $m$, and MSE is chosen as the loss function for TTE. Then the objective function can be defined as 
\begin{align}\label{equ:obj_1}
	\mathcal{L}_{\theta_b}=\sum_{i}^{n_m}\left\|y_i-\widehat{y_i}\right\|^{2}
\end{align}
where $n_m$, $y$ and $\Tilde{y}$ denotes the number of trajectory for client $m$, the ground truth and estimated value respectively. $\theta_b$ are all trainable parameters in the base model. In this study, the  stochastic gradient descent
(SGD) algorithm as the default optimizer for both base and personalized models, and the FedAvg \cite{mcmahan2017communication} is  set as  federated learning strategy. However, another gordian knot is that there is an unequal amount of data between users, since each driver receives different taxi orders and travels different distances every day (\figref{fig:driver_profiles}). We here utilize a simple method mentioned in \cite{mcmahan2017communication} that the global model $F^{t+1}$ is equal to the weighted-average sum of all clients: $F^{t+1} \leftarrow \sum_{m=1}^{M} \frac{n_{m}}{n} f_{m}^{t}$, where $f_{k}^{t}$ is the local model, $\alpha$ is the learning rate, $M$ is the amount of selected clients and each client owns $n_{m}$ samples at $t$ round, and $n$ is the total number of samples for all selected clients at $t$ round.
After that, we freeze the localized global model and train the fine-tuned personalized model to learn each client/user's personal driving habits, which is to fit the residual error in \eqnref{equ:obj_1} as 
\begin{align}\label{equ:obj_2}
	\mathcal{L}_{\theta_p}&=\sum_{i}^{n_m}\left\|y_i-\Tilde{y_i}\right\|^{2},\\
	&=\sum_{i}^{n_m}\left\|y_i-\widehat{y_i}-\widehat{bias}\right\|^{2},
\end{align}
where $\theta_p$ are all trainable parameters in the personalized model, and $\Tilde{y}=\widehat{y}-\widehat{bias}$ is the final output of our proposed model (as is shown in \figref{fig:model}). In the final, to protect the client's privacy risk of model weights, we construct the noise based on differential privacy and add them into trained localized global model for each client before uploading them to the cloud server. Then the cloud server can acquire an aggregate global model and conduct the next round of federated learning. Here, our federated training procedure at time step $t_k$ can be shown in the Algorithm \ref{alg:alg1}. It is noted that we do not design complicated structure for the personalized model to avoid the potential overfitting on limited personal trajectory data. 

\begin{algorithm}[h]
	\caption{Model Training procedure}\label{alg:alg1}
	\LinesNumbered
	\textbf{Global parameters: } $F^{t}$ is the global model at $t$ round; $\alpha$ is the base model learning rate; $M$ is the amount of selected clients. \\
	\textbf{Local parameters: }$f_m$ is the localized global model for client $m$;  $p_m$ is the personalized model for client $m$, \textit{epoch} is the number of local epoch; $\mathcal{D}_m$ is the personal privacy trajectory data on  local device; $\varepsilon$ is the parameter of differential privacy.\\
	\textbf{Server: }\\
	initialize $F^{0}$\\
	\For{ \textit{each round} $t \in \{ 1,2, ...\}$}{
		$I_t \leftarrow$ random set of $M$ client's; \\
		\For{each client $m \in I_t$ \textbf{in parallel}}{
		$f_m \leftarrow$ \textbf{ClientUpdate}($F^{t-1}$, \textit{$epoch$}, $\alpha$);}
		$F^{t+1} \leftarrow \sum_{m=1}^{M} \frac{n_{m}}{n} f_{m}^{t}$
	}
	\textbf{Client Update: }\\
	\For{$i \in \{ 1,2,..., \textit{$epoch$} \}$}{
		$f_m \leftarrow f_m- \alpha \nabla \mathcal{L}_{\theta_b}(f; \mathcal{D}_m)$ \qquad // $\nabla\mathcal{L}_{\theta_b}(f; \mathcal{P}_m)$ obtained from \eqnref{equ:obj_1}
	}
	$f_m \stackrel{+n o i s e}{\longleftarrow} f_m $ \qquad //  construct noise with differential privacy $\varepsilon$ \\ 
	\textbf{return} $f_m$ \\
	\textbf{Fine-tune} personalized model $p_m$ based on $\mathcal{L}_{\theta_p}$ for each client $m$ before inference.
	
\end{algorithm}

\label{sec:model}
\section{Experiments}
\label{sec:exp_}


\begin{table*}[tb]
	\centering
	\caption{Overall performance comparison under two public datasets. Here, "+Personalized" denotes the training procedure of localized global model and personalized model described in Sec.\ref{model_training}} 
	\vskip -0.1in
	\setlength{\tabcolsep}{1.0mm}{	\begin{tabular}{l lc c c |c c c }
			\toprule \multirow{2}{*}{Training Strategy}&  \multirow{2}{*}{Models}&\multicolumn{3}{c}{Chengdu} & \multicolumn{3}{c}{Xi'an} \\
			\cline{3-5}  \cline{6-8}  
			&& {\small MAE (sec)} & {\small RMSE (sec)} & {\small MAPE (\%)} & {\small MAE (sec)} & {\small RMSE (sec)} & {\small MAPE (\%)} \\
			\midrule
			\multirow{4}{*}{Centralization}
			&T-GCN&201.28& 297.59&  12.09&225.89& 361.91&  14.50 \\
			&ConSTGAT&199.62&295.53& 11.78&187.57& 295.33&  12.03 \\
			&DeepTTE&194.47&293.15&12.36&167.85&266.26&11.28 \\
			&WDR & 231.22&355.76&15.88&207.52&304.63&14.46 \\
			&MURAT&267.83&394.29&17.78&225.39&353.16&14.37\\
			&DeepGTT&251.71&369.26&18.12&247.83&367.25&15.09\\
			\hline
			\multirow{3}{*}{Fed-Avg}
			&T-GCN&281.45& 393.61&  17.25&305.68& 457.92&  19.34 \\
			&ConSTGAT&279.14& 390.87&  16.82&253.82& 373.67&  16.04\\
			&DeepTTE&271.93&387.73&17.64&227.14&336.89&15.04 \\
			&WDR &353.29&469.52&24.56&291.96&404.72&19.29 \\
			&MURAT&369.17&458.16&24.39&311.85&427.68&21.24 \\
			\hline
			Fed-Avg&\multirow{4}{*}{Localized Global Model}&285.53&403.36&18.21&238.37&368.16&15.57\\
			Fed-PA&&291.98&411.46&18.27&234.26&356.84&15.69\\
			Fed-Ensemble&&272.43&393.94&17.28&229.37&331.41&14.93\\
			 +Personalized (ours) &&\textbf{237.46}&\textbf{360.38}&\textbf{17.11}&\textbf{209.32}&\textbf{311.59}&\textbf{14.28}\\
			
			\bottomrule
	\end{tabular}}
	\label{table:result}
\end{table*}

\begin{figure*}[t]
	\centering
	\includegraphics[width=1\linewidth]{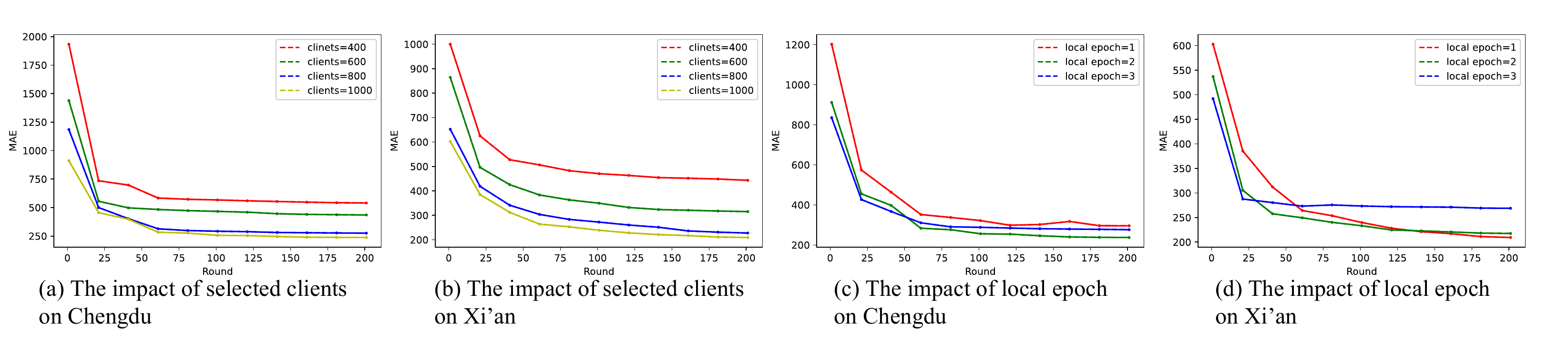}
	\caption{The effects of the number of selected clients and the local training epoch.} 
	\label{fig:hyper}
\end{figure*}

\subsection{Baselines}

In terms of travel time estimation models, we compare our proposed methods with baseline models including DeepTTE, WDR, MURAT, DeepGTT, T-GCN, and ConSTGAT in different training strategies (e.g. Fed-Avg and centralized training).

\begin{itemize}
	\item[$\bullet$] \textbf{DeepTTE} \cite{wang2018will} is an end-to-end deep learning framework, which infers the trajectory travel time of both the entire path and each local path simultaneously.

	 \item[$\bullet$] \textbf{WDR} \cite{wang2018learning} predicts the travel time along a given route at a specific departure time, and then jointly trains wide linear models, and recurrent neural networks together to take full advantage of all three kinds of models.
	 
	\item[$\bullet$] \textbf{MURAT} \cite{li2018multi} is a multi-task representation learning method by utilizing the underlying
	road network and the spatio-temporal prior knowledge.
	
	\item[$\bullet$] \textbf{DeepGTT} \cite{li2019learning} learns the travel time distribution through the deep generative model, which takes the real-time traffic condition into consideration.
	
	\item[$\bullet$] \textbf{T-GCN} \cite{zhao2019t} is a temporal graphical model that combines the GCN and GRU to simultaneously characterize the spatio-temporal dependencies.
	
	\item[$\bullet$] \textbf{ConSTGAT} \cite{fang2020constgat} adopts a graph attention mechanism to explore the joint relations of spatio-temporal information.
\end{itemize}
All baseline models but DeepGTT (due to structural reason) are implemented in both centralized training and federated learning strategies. For centralized training, data is shared by different users, and the model trained with shared data succeeds in capturing the universal mobility patterns and achieves better modelling performance. However, privacy issues are completely ignored. And in federated learning, the model is directly trained and executed in personal devices with only local private data. Without
sharing users' private data, this training strategy protects the personal privacy but fails to provide competitive performance. In this paper, we also apply three federated learning strategies to our proposed system. Among them, Fed-PA and Fed-Ensemble are recent personalized federated learning strategies for urban computing.
\begin{itemize}
	\item[$\bullet$] \textbf{Fed-Avg} \cite{mcmahan2017communication} In Federated Averaging,  a subset of the total devices are selected and trained locally for E number of epochs at each round, and then the updates of resulting local model are averaged. 

	 \item[$\bullet$] \textbf{Fed-PA} \cite{feng2020pmf} Feng et al. proposed a personal adaptor to be fine-tuned in the local device for better personal pattern modelling. In our experiment, the time embedding layer is filtered by a trainable same-size vector with sigmoid function using the multiply operation.

	\item[$\bullet$] \textbf{Fed-Ensemble} \cite{liu2020privacy} Yu et al. proposed an ensemble clustering-based scheme for traffic flow prediction by grouping organizations/users into clusters before applying the Fed-Avg algorithm. In our experiment, we use the average locations of every user's trajectories to cluster users and then implement federated averaging in the corresponding clusters to get the global model set. Finally, the global model is generated by the ensemble learning scheme.

\end{itemize}
\subsection{Metrics}
 We evaluate the performance of travel time estimation with RMSE (Root Mean Square Error),
MAE (Mean Absolute Error), and MAPE (Mean Absolute Percentage Error).
\begin{align*}
	R M S E&=\sqrt{\frac{1}{n} \sum_{i}^{n}\left\|\Tilde{y}_{i}-y_i\right\|^{2}}, \\
	M A E&=\frac{1}{n} \sum_{i}^{n}\mid\Tilde{y}_{i}-y_i\mid,\\
	M A P E&=\frac{1}{n} \sum_{i}^{n}\mid\frac{\Tilde{y}_{i}-y_i}{y_i}\mid,
\end{align*}
where $n$, $y$ and $\Tilde{y}$ denotes the number of trajectory samples, the ground truth and estimated value respectively.
\subsection{Experimental Setup}
 The experiments are implemented with PyTorch 1.6.0 and Python 3.6 and the models are trained with a RTX2080 GPU on Ubuntu 16.04 OS platform. We trained the models using SGD optimizer with an initial learning rate of 0.001 on both Chengdu and Xi'an datasets, and early stopping is applied to avoid overfitting. For the hyper-parameters of federated training, we set clients=1000, local epoch=2 in several federated learning strategies, \textcolor{black}{ except for our proposed training strategy \textit{Base + Personalized}. This is because the local training of our prediction model is a two-step training process. Here, we implement local epoch=2 to train the localized global model and local epoch=1 to fine-tuned personalized model.}
 
 \begin{table*}[tb]
	\centering
	\caption{\textcolor{black}{Overall simulation performance comparison of our proposed method and baseline strategies for travel time estimation (on Chengdu and  Xi'an Dataset) under two scenarios, including both non-rush and rush hours.}} 
	\renewcommand\arraystretch{1.5}
	\setlength{\tabcolsep}{1.0mm}{	\begin{tabular}{c l lc c c |c c c }
			\toprule 			\multirow{2}{*}{Data}& \multirow{2}{*}{Training Strategy}&  \multirow{2}{*}{Models}&\multicolumn{3}{c}{Non-Rush Hours} & \multicolumn{3}{c}{Rush Hours} \\
			\cline{4-6}  \cline{7-9}  
			\multirow{7}{*}{\rotatebox[origin=c]{90}{Chengdu}}&&& {\small MAE (sec)} & {\small RMSE (sec)} & {\small MAPE (\%)} & {\small MAE (sec)} & {\small RMSE (sec)} & {\small MAPE (\%)} \\
			\midrule
			
			&Fed-Avg&\multirow{4}{*}{Localized Global Model}&314.20&405.26&18.93&347.15&488.16&20.78\\
			&Fed-PA&&323.39&413.33&19.24&364.77&507.82&21.04\\
			&Fed-Ensemble&&307.27&403.65&18.91&321.32&463.59&19.23\\
			&+Personalized (ours) &&\textbf{256.92}&\textbf{384.18}&\textbf{16.17}&\textbf{291.46}&\textbf{410.33}&\textbf{18.75}\\
			\midrule
			\multirow{4}{*}{\rotatebox[origin=c]{90}{Xi'an}}&
			Fed-Avg&\multirow{4}{*}{Localized Global Model}			&{262.275}  &{369.851}  &{16.20}   &{289.782}  &{445.517}  &{17.78}  \\
			&Fed-PA&												&{269.947}  &{377.217}  &{16.47}   &{304.492}  &{463.461}  &{18.00}  \\
			&Fed-Ensemble&											&{256.489}  &{368.381}  &{16.18}   &{268.219}  &{423.091}  &{16.46}  \\
			&+Personalized (ours) &									&\textbf{214.455}  &\textbf{350.611}  &\textbf{13.85}   &\textbf{243.290}  &\textbf{374.479}  &\textbf{16.01}  \\

			\bottomrule
	\end{tabular}}
	\label{table:result_simulation}
\end{table*}

\subsection{Experimental Results}
This section presents the performance comparison with baseline models under different training strategies, including centralized training and several federated learning strategies. Then we analyze the effect of hyper-parameters in the training procedure of our proposed travel time estimation system. Especially, we also conducted a real-world case study in Xi'an, which visualizes the learned global state of our proposed system on the road network.

\subsubsection{Overall Performance} We compare our proposed methods with different estimation models and training strategy combinations under two real-world datasets. These can be divided into privacy-preserving models under federated training and privacy-leakage models under centralized training. \tableref{table:result} shows the performance of travel time estimation. From these results, we first find that centralized training usually achieves better performance than federated training regarding the same prediction model. On the other hand, the best performances are both achieved by privacy-leakage DeepTTE. Furthermore, we can conclude that our proposed method is superior to other privacy-preserving models. First, our global model combined with the personalized model achieves the best performance among privacy-preserving models with an improvement of 12.67\% and 7.85\% under two datasets, even though it still has competitive performance compared with the best privacy-leakage models. Second, our federated training procedure also performs better than other federated learning strategies, such as Fed-PA and Fed-Ensemble. Besides, only the global model with Fed-Avg has achieved near the performance of DeepTTE with Fed-Avg. This means that the trained traffic states of local drivers have extreme reliability. In summary, our proposed methods achieve promising performance while protecting personal privacy in federated training.

\textcolor{black}{\subsubsection{Simulation Performance Under Different Scenarios} The simulation experiments have been conducted to verify the correctness and usefulness of our online TTE framework. The corresponding result comparisons are shown in \tableref{table:result_simulation}. To be clear, we here present the model performance in non-rush hours (0:00-7:00, 9:00-17:00, and 19:00-24:00) and rush hours (7:00-9:00 and 17:00-19:00), respectively. As we have heard, since traffic state tends to be congested during the rush hours, it is a challenge to accurately estimate travel time. We take an example for Chengdu datasets, our proposed framework that integrates with the personalized model, still improves at least 15\%, 16\%, and  8\%  in RMSE, MAE, and MAPE. In terms of non-rush hours, our model obtains at least 18\%,  6\%, and 15\% improvement in these three metrics. We can conclude that our model significantly outperforms overall federated strategy and TTE baselines, demonstrating that the generalization of our proposed framework can better adapt to different scenarios.}

\subsubsection{The Effects of Hyper-parameters on Federated Learning}
In the experiment, we first analyze the effects of the number of selected clients. In each training round of the proposed system, we only choose a fraction of clients to participate in the current optimization procedure. \figref{fig:hyper} (a) and \figref{fig:hyper} (b) present the effects of the number of selected clients on the final performance on two datasets. Generally, we can observe that the final performance is improved continuously, with the number of clients increasing from 400 to 1000 for both datasets. This indicates that the generated global traffic states need large amounts of users' trajectories to train. We also study the effects of the local training epoch of the localized global model on the system performance in \figref{fig:hyper}(c) and \figref{fig:hyper}(d). We find that local epoch=2 and local epoch=1 are the optimal settings for Xi'an and Chengdu datasets, respectively. This is because Xi'an's road network is simpler than Chengdu, and the training process for the Xi'an dataset needs fewer iterations than Chengdu.

\begin{figure}[!b]
	\centering
	\setlength{\belowcaptionskip}{-0.3cm}
	\includegraphics[width=1\linewidth]{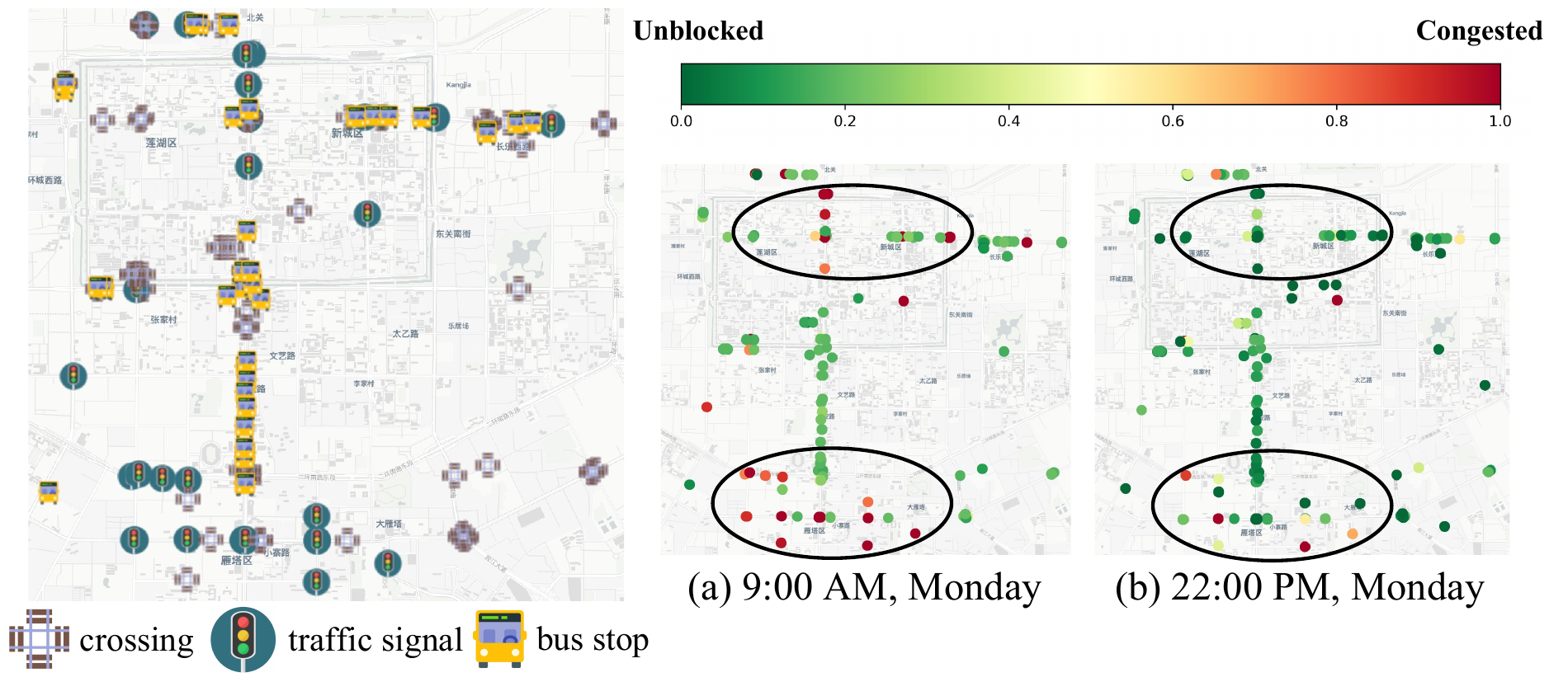}
	\caption{The time-consuming ratio for some nodes in the Xi'an road network. Here, we select three types of nodes including "crossing", "traffic signal" and "bus stop".}
	\label{fig:node}
\end{figure}
\begin{figure*}[!bht]
	\centering
	\setlength{\belowcaptionskip}{-0.1cm}
	\includegraphics[width=0.85\linewidth]{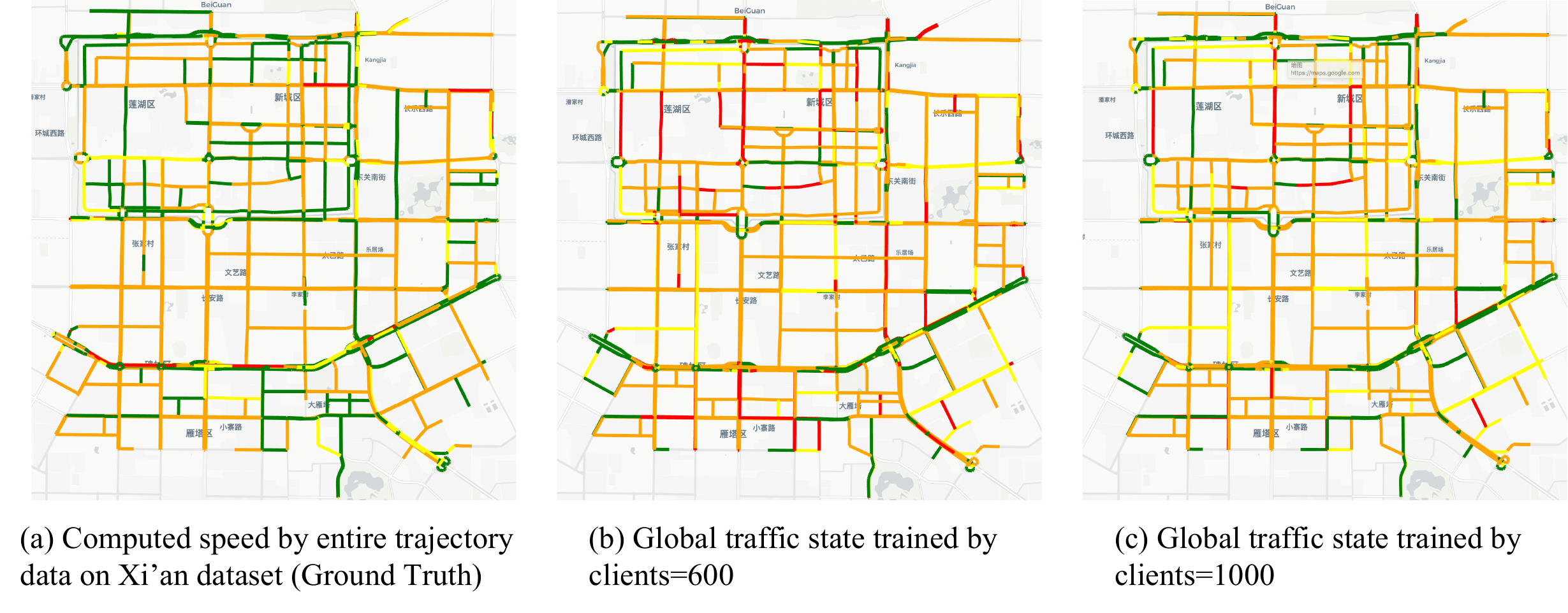}
	\caption{The aggregated global traffic state in Xi'an. Here we use four kinds of colors to represent the different road states, which can be defined as: 1) red - very congested, 2) orange - congested, 3) yellow - slow, 4) green - unblocked.}
	\label{fig:road_network}
\end{figure*}

\subsubsection{Case Study}

Our proposed system can not only conduct path travel time estimation for personal trips, but also learn the global traffic state of the links and nodes. 

On the one hand, we first provide the aggregated global traffic state for different nodes, which are depicted in \figref{fig:node}. On the other hand, we select three types of nodes in a road network and use the time-consuming ratio to represent congestion status. The time-consuming ratio is calculated by dividing the corresponding mean value of learned travel time distributions by the maximum mean value among all nodes (note that we have filtered out the top 1\% most enormous node travel time). From \figref{fig:node}, we can find that nodes with the "traffic signal" type are more time-consuming than the other two types of nodes, and most nodes in the morning peak are easy to become congested. These show that our node travel time estimation is reasonable in both spatial and temporal aspects.

On the other hand, we depict the aggregated global states of links and compare them with the ground truth computed by the original taxi trajectories. Especially, we compare the aggregated road conditions with two different numbers of selected clients (clients=600 and clients=1000). We mark it with the unblocked state for the road segments without taxi trajectories. Since the speed limit varies from road to road, primarily defined by road type or road length, we use four colors to represent the different road states (very congested, congested, slow, and unblocked). We divide the limiting-velocity for each road type equally. For example, the rate-limiting of the primary road is $60kph$, so the interval between very congested is $[0, 15)$, congested is $[15, 30)$, slow is $[30, 45)$, unblocked is $[45, 60)$. The compared result is shown in \figref{fig:road_network}. Comparing the aggregated road condition and ground truth, the generated traffic states under the selected client's two settings are similar to the ground truth. Most road segments have the same road states. Furthermore, we find that the more clients participate in the federated aggregation process, the more accurate the global traffic state will be.

\begin{figure*}[t]
	\centering    
	\includegraphics[width=1\linewidth]{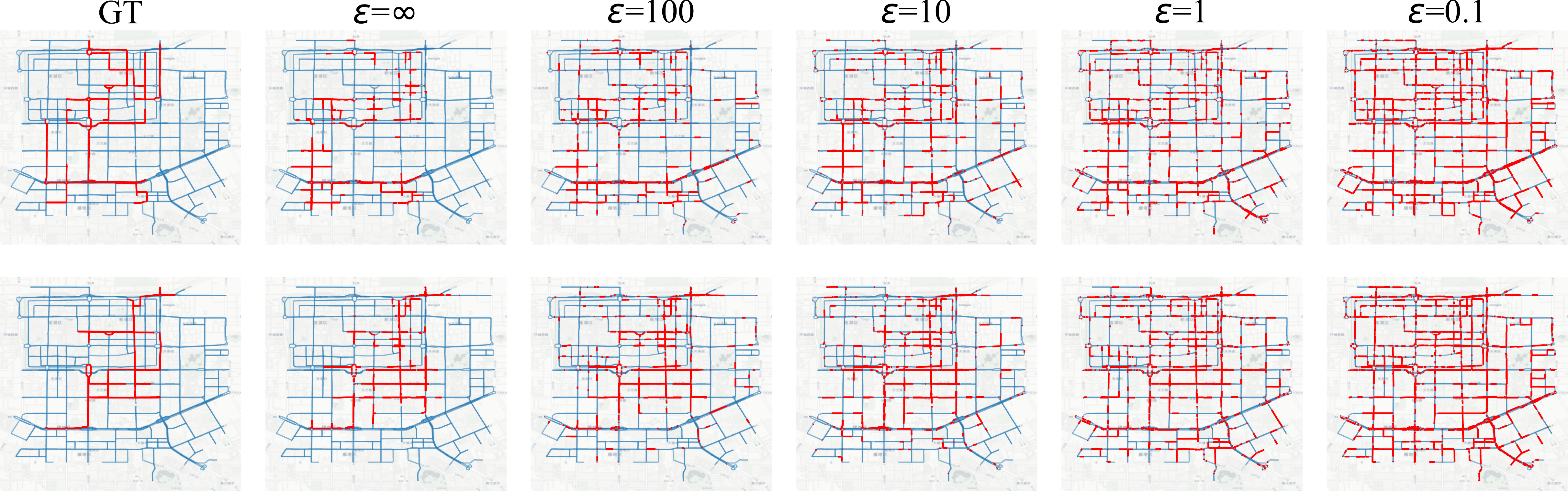}   
	
	\caption{The effectiveness of attack risk with changing differential privacy parameter $\varepsilon$. }
	\label{fig:qualitative_attack}     
\end{figure*}

\begin{figure}[h]\setcounter{subfigure}{0}
	\centering    
	\subfigure[Attack results on Xi’an] {
		\includegraphics[width=0.48\linewidth]{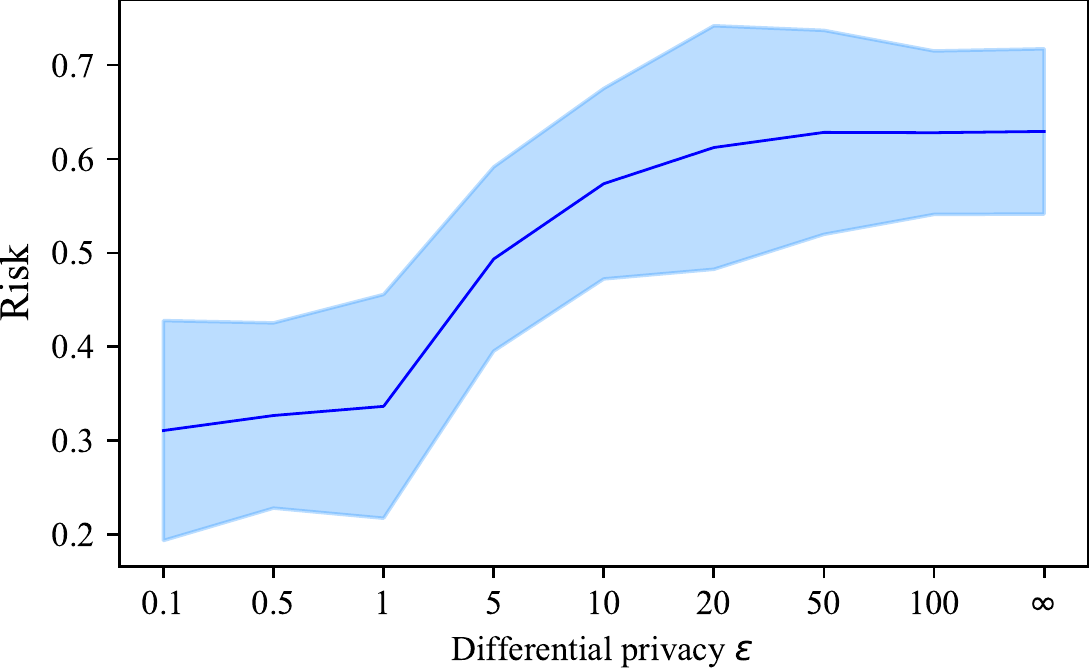}}     
	\hfill
	\subfigure[Attack results on Chengdu] {
		\includegraphics[width=0.48\linewidth]{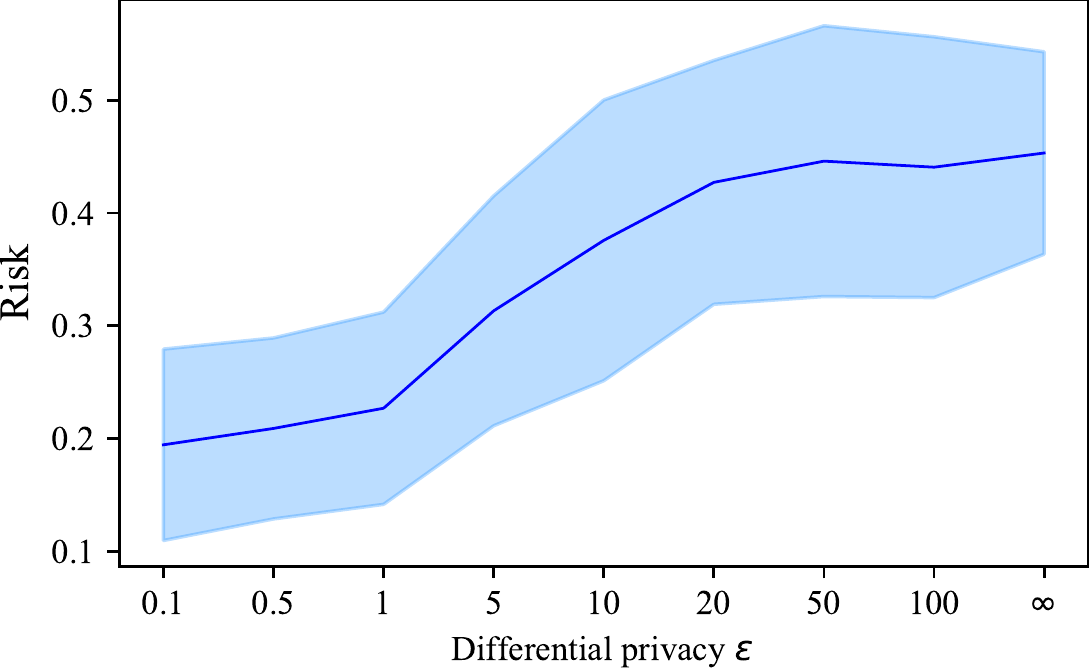}}     
	\caption{The privacy attack result with different differential privacy parameter $\varepsilon$.}\label{fig:quantitative_attack}
\end{figure}

\subsection{Privacy Risk Analysis}
\textcolor{black}{As the introduction about analyzing the privacy risk of uploading the base model in \secref{sec:privacy_sec}, we conduct the privacy risk analysis for our proposed framework, and show the effectiveness of the protection on local devices. Here, we have two objectives of privacy analysis. One is to validate our local data protection method based on our differential privacy mechanism; the other is to evaluate the effects of differential privacy on the prediction accuracy of travel time.}    
\subsubsection{Privacy-preserving Performance}
To evaluate the effectiveness of our privacy protection method, we define the  privacy $risk = \frac{|\Omega_i \cap \tilde{\Omega_i}|}{|\Omega_i|}$ metrics to measure the privacy-preserving performance with different parameter settings of $\varepsilon$ for differential privacy, where $\Omega_i$  is the set of personal real travelled road segments and $\tilde{\Omega}_i$ is the set of road segments revealed by privacy attack.

\textcolor{black}{In this study, we measure the privacy-preserving performance from two aspects: a qualitative and quantitative analysis. As shown in \figref{fig:qualitative_attack}, we compare the effectiveness of attack risk with different setting of differential privacy parameter $\varepsilon$, which is the same case with \figref{fig:privacy_attacks}. We can observe that the attack risk can be reduced significantly by utilizing differential privacy mechanisms. Furthermore, we can also see that from left to right, the noise intensifies, and the risk (overlap) declines dramatically. As \figref{fig:qualitative_attack} shows, it is difficult to attack mobile users' location privacy when the $\varepsilon$ reduces to 10. Moreover, when $\varepsilon$ is lower than 10, the privacy attack would have unobservable effects, which ensures the user's privacy.. In addition, we conduct the quantitative analysis of privacy risk, and \figref{fig:quantitative_attack} represents the average risk of overall clients/users in federated training procedure for both datasets, respectively. As \figref{fig:quantitative_attack} shows, we can also know that the clients are exposed to high risk when the strength of noise is small. Especially, the trend of privacy risk for Xi'an, which is shown in \figref{fig:quantitative_attack} (a), validates our conclusions.}

\subsubsection{The Effects of Differential Privacy on Prediction Performance} \textcolor{black}{To efficiently protect location-based privacy on the local devices, we expect to utilize differential privacy mechanism in the ubiquitous computing system. However, greater noise always leads to the limited performance of the estimation task. Thus, how to balance the privacy protection and better performance becomes an important topic in ubiquitous computing. \tableref{tab:privacy_tab} presents the effects of differential privacy parameter $\varepsilon$ on the travel time estimation performance under both two datasets.  We further present the relative drops of estimation accuracy based on the  performance without differential privacy ($\varepsilon=\infty$). We can observe that in terms of MAE, the performance in Chengdu and Xi'an gradually
 decreases when $\varepsilon$ reduces from 100 to 10. In Chengdu dataset, the model performance is relatively sensitive when $\varepsilon$ reduces from 100 to (1,0.1,0.01), compared with the Xi'an dataset. We suppose this may rely on the different road network structure between two cities. To investigate that, we here also depict the attack risk with changing differential privacy parameter $\varepsilon$ in \figref{fig:qualitative_attack}. } We can find that the performance of Chengdu datasets that has relatively complex road network structure has low attack risk as \figref{fig:quantitative_attack} (b). For Xi'an dataset, the varying $\varepsilon$ has relatively small influence on prediction performance when $\varepsilon$ reduces from 100 to 1, although it has higher attack risk. In summary, our method can achieve better trade-off between the model performance and privacy risk of difference attack, and these ensure the availability of our proposed privacy-preserving mechanism. 

\begin{table}[!h]
	\centering
	\renewcommand\arraystretch{1.5}
	\caption{MAE comparison of our method with changing differential privacy parameter $\varepsilon$.}
	\label{tab:privacy_tab}
	\begin{tabular}{c|cccccc}
		\toprule
		Dataset$\backslash$$\varepsilon$ & 0.01&0.1&1&10&100&$\infty$\\
		\hline
		Chengdu&367.51 &353.24&338.22&297.90&253.15&237.46\\
		Drops (\%)& 54.77& 48.76& 42.43& 25.45&  6.61 &  0\\
		Xi'an& 291.47&284.38&255.54&233.15&213.67&209.32\\
		Drops (\%)& 39.25& 35.86& 22.08& 11.38&  2.09&  0\\
		\bottomrule
	\end{tabular}
\end{table}\label{sec:exp}
\section{Related Work}
\label{sec:related}
\subsection{Travel Time Estimation}
Existing approaches of travel time estimation can be classified into four groups: road segment based, path based, deep learning based and graph neural network based methods.

\textbf{Road Segment Based and Path Based Methods.}
The road segment based methods use the speed data sampled from loop detectors to infer each individual road segment's travel time independently \cite{rice2004simple} but ignore the correlations between the
road segments, such as the factors of traffic lights, and left/right turns. Path based approaches have been proposed to resolve those issues, which can roughly be divided into two types: 1) nearest neighbor search \cite{tiesyte2008similarity,wang2019simple}, which estimates the travel time by averaging the historical trajectories' travel time. 2) trajectory regression methods \cite{ide2011trajectory}, which predicts the travel time of road segments via public features, for instance, road types, road lanes, road length, etc. 

\textcolor{black}{\textbf{Deep Learning Based Methods.} With the development of deep learning in these years, various studies have been designed to enhance the performance of travel time estimation. For instance, DeepTTE \cite{wang2018will} presents an end-to-end framework that predicts the travel time of the whole path directly by using geo-convolution on the GPS
sequences to capture spatial features. \textit{Li et al.} \cite{li2019learning} designs a deep generative model
to learn the travel time distribution for each road segment using grid-based traffic conditions and basic road network features. \textit{Wang et al.} \cite{wang2018learning} models the problem of travel time estimation as a spatio-temporal regression problem by a Wide-Deep-Recurrent learning model \cite{cheng2016wide}, and this work considers various features including spatial information (the characteristics of route, such as the road segment and traffic light information), temporal information (such as a month and a day, the holiday indicator), traffic information (such as estimated road condition) and personalized information (such as driver profile).
}

\textbf{Graph Neural Network Based Methods.} Very recently, the
advances of graph neural network have promoted a myriad of work in travel time estimation problem. For example, ConSTGAT \cite{fang2020constgat} employs a graph attention
mechanism onto the spatial-temporal features by integrating traffic  and contextual information, \textcolor{black}{in which the input feature includes the road segment-based information, historical road condition and departure time of each trajectory.} Deepist \cite{fu2019deepist} enhances the  power of convolutional neural network to capture moving behaviors embedded in paths by introducing the image of urban in grid level. \textcolor{black}{GraphTTE \cite{wang2021graphtte} designs a Multi-layer Spatio-temporal Graph frame, which consists of static and dynamic networks. In particular, the dynamic networks implement the GCN and gate recurrent unit to model the historical traffic  characteristics, and the static network employs GCN to model the basic road attributes. Especially, \textit{Hong et al.} \cite{hong2020heteta} transforms the road network into a multi-relational network and introduce a vehicle-trajectories based network to jointly consider the traffic behavior pattern (traffic speed, volume).}

\textcolor{black}{However, those graph-based approaches ignore the complex relations between road segments and intersections from road network. Meanwhile, their methods need to model the real traffic dynamics, which is difficult to acquire due to the power and communication limitations of mobile devices. In this
paper, we model the adjacencies in both node-wise graph and link-wise graph simultaneously. And the online traffic state can be obtained via our aggregated global model.}
\subsection{Privacy Protection Methods}
We here discuss the related work from two directions: 1) user data protection mechanism, and 2) federated Learning.

\textbf{User Data Protection Mechanisms.} a series of techniques aiming to protect the user trajectories have been proposed, which can be classified with two categories \cite{zheng2015trajectory}: 1) real-time continuous location-based services, and 2) publication of historical trajectories. The former intends to protect the scenario, for example, asking the Apps to give the traffic
conditions that are 1 km around me. Multiple techniques have been proposed including spatial cloaking \cite{mokbel2006new}, mix zones \cite{beresford2003location} and path confusion \cite{hoh2010achieving}. The latter schemes to protect the trajectories, and its related works include clustering-based \cite{abul2008never}, generalization-based \cite{nergiz2008towards},  suppression-based \cite{terrovitis2008privacy}, and grid-based \cite{gidofalvi2007privacy} approaches.

\textbf{Federated Learning.} It is a mechanism to coordinate the joint training model of multiple participants, which is emerging in the era of big data and the development of artificial intelligence technology \cite{li2020federated}. Meanwhile, communication cost and communication efficiency have become one of the key bottlenecks of federated learning. A bunch of work has been proposed to analyze how to reduce the communication cost of federated learning \cite{mcmahan2017communication,li2018federated}.  Further, federated learning does not exchange data directly and has a higher privacy guarantee than traditional centralized machine learning training, but federated learning itself does not provide comprehensive and sufficient privacy protection. 
Recently, researchers have applied differential privacy operation \cite{dwork2006differential} on the whole model in the aggregation stage of federated learning \cite{geyer2017differentially}.  

\textbf{Personalized Federated Learning.} Personalized federated learning \cite{vanhaesebrouck2017decentralized} differs from traditional federated learning in that it does not require all participants to end up using the same model, but allows each participant to fine-tune the model based on their own data to generate their own unique personalized model. Models often perform better on local test sets after personalized tweaks. A bunch of work has been proposed from meta-learning \cite{fallah2020personalized}, multi-task learning \cite{dinh2021fedu}, hypernetwork \cite{shamsian2021personalized}, knowledge distillation \cite{li2019fedmd} to personalized model \cite{collins2021exploiting}. For example, the goal of meta-learning \cite{fallah2020personalized} is to train highly adaptive models that can be trained to solve new tasks in a small number of samples.And in multitasking learning \cite{dinh2021fedu}, models among the clients learn together by taking advantage of commonalities and differences between tasks.

\textcolor{black}{In this paper, our proposed federated learning system can be viewed as a case of
\textit{base model + personalized model}, which estimates the travel time by fusing the 
outputs of both localized global and personalized model. Moreover, our aggregated global model can generate real-time road condition, which plays an important role of the intelligent transportation system, compared with single fusion of the previous personalized model strategies.}

\label{sec:relate}
\section{Conclusion}

\textcolor{black}{In allusion to the TTE problem, this paper proposes an online personalized federated learning framework to fill the gap between personal privacy and model performance by dynamically updating the global traffic state. Specifically, we estimate the travel time through the training strategy of the \textit{base model + personalized model}, where the base model serves to produce the unbiased global traffic state for all clients, and the latter is only trained locally to act on fitting the personalized driving habits through the profile features. We evaluate the effectiveness of our proposed framework from three aspects: 1) overall performance compared with state-of-the-art in travel time estimation models and federated learning strategies; 2) a case study depicting the global state from links and nodes, respectively, to show the aggregation performance of our system. 3) privacy risk analysis from both quantitative and qualitative views, respectively, to prove the effectiveness of our privacy-preserving mechanism.}

\textcolor{black}{In the future, we will further test the performance with more federated learning strategies, such as clustering the personal driving features together, since the drivers' habits have multiple similarities. Meanwhile, we will also reduce the communication cost since the online algorithm frequently needs to upload and download model weight between clients and the cloud server.}

\section{Acknowledgment}

This work was partially supported by National Key Research and Development Project (2021YFB1714400) of China and  Guangdong Provincial Key Laboratory (2020B121201001).

\label{sec:conclude}




\bibliographystyle{unsrt}
\bibliography{main}

\end{document}